\documentclass[10pt,twocolumn,letterpaper]{article}

\usepackage[pagenumbers]{cvpr} 

\usepackage{multirow}
\usepackage{tabularx}
\usepackage{tablefootnote}
\usepackage{colortbl}
\usepackage{amsmath}
\usepackage{footnote}
\makesavenoteenv{figure}
\usepackage{float}
\usepackage[table,x11names,dvipsnames]{xcolor}

\usepackage{textcomp}

\usepackage{booktabs} 
\usepackage{array} 
\usepackage{amssymb} 
\newcommand{\cmark}{\checkmark}
\newcommand{\xmark}{\texttimes}
\usepackage{graphicx}
\usepackage{textcomp}
 \newcommand{\mytexttildenew}{\raisebox{0.5ex}{\texttildelow}}

\definecolor{purple}{HTML}{9929BD}
\definecolor{light_green}{HTML}{669C35}

\definecolor{cvprblue}{rgb}{0.21,0.49,0.74}
\usepackage[pagebackref,breaklinks,colorlinks,allcolors=cvprblue]{hyperref}

\def\paperID{*****} 
\def\confName{CVPR}
\def\confYear{2026}

\title{StreamReady: Learning \emph{What} to Answer and \emph{When} in Long Streaming Videos
\vspace{-.5cm}}

\author{Shehreen Azad\textsuperscript{1}  \qquad  \qquad Vibhav Vineet\textsuperscript{2} \vspace{3pt} \qquad \qquad Yogesh Singh Rawat\textsuperscript{1} \\
\textsuperscript{1}Center for Research in Computer Vision, University of Central Florida; \qquad \textsuperscript{2}Microsoft Research\\
\url{https://sacrcv.github.io/StreamReady-website/}} 

\begin{document}

\twocolumn[{%
\renewcommand\twocolumn[1][]{#1}%
\maketitle
\begin{center}
    \centering
    \captionsetup{type=figure}
    \vspace{-10pt}
    \includegraphics[width=.59\linewidth]{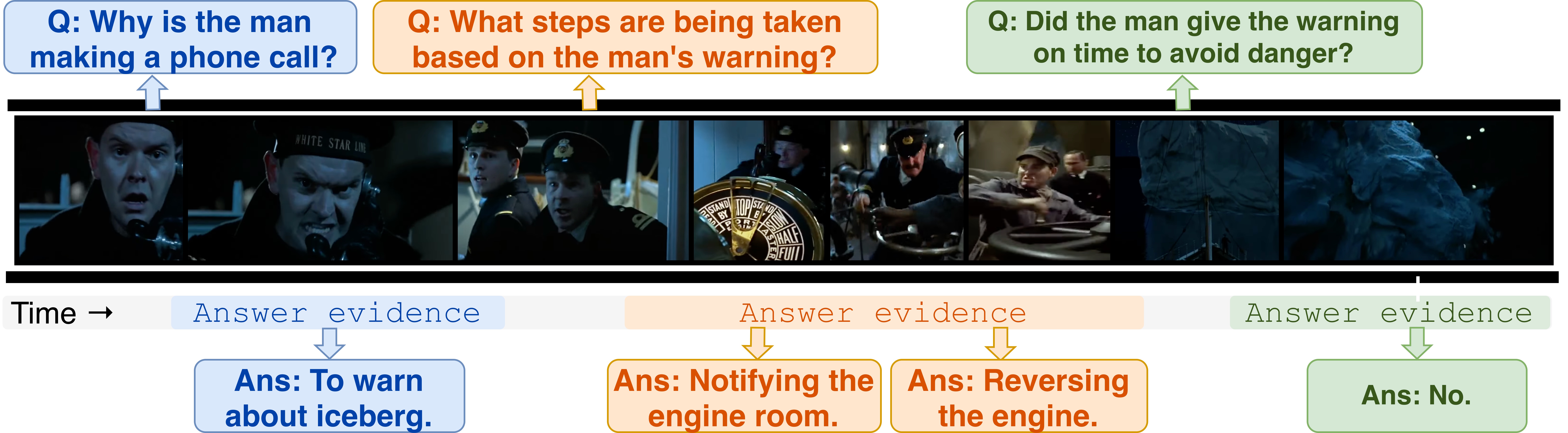}
    \hfill
    \includegraphics[width=.39\linewidth]{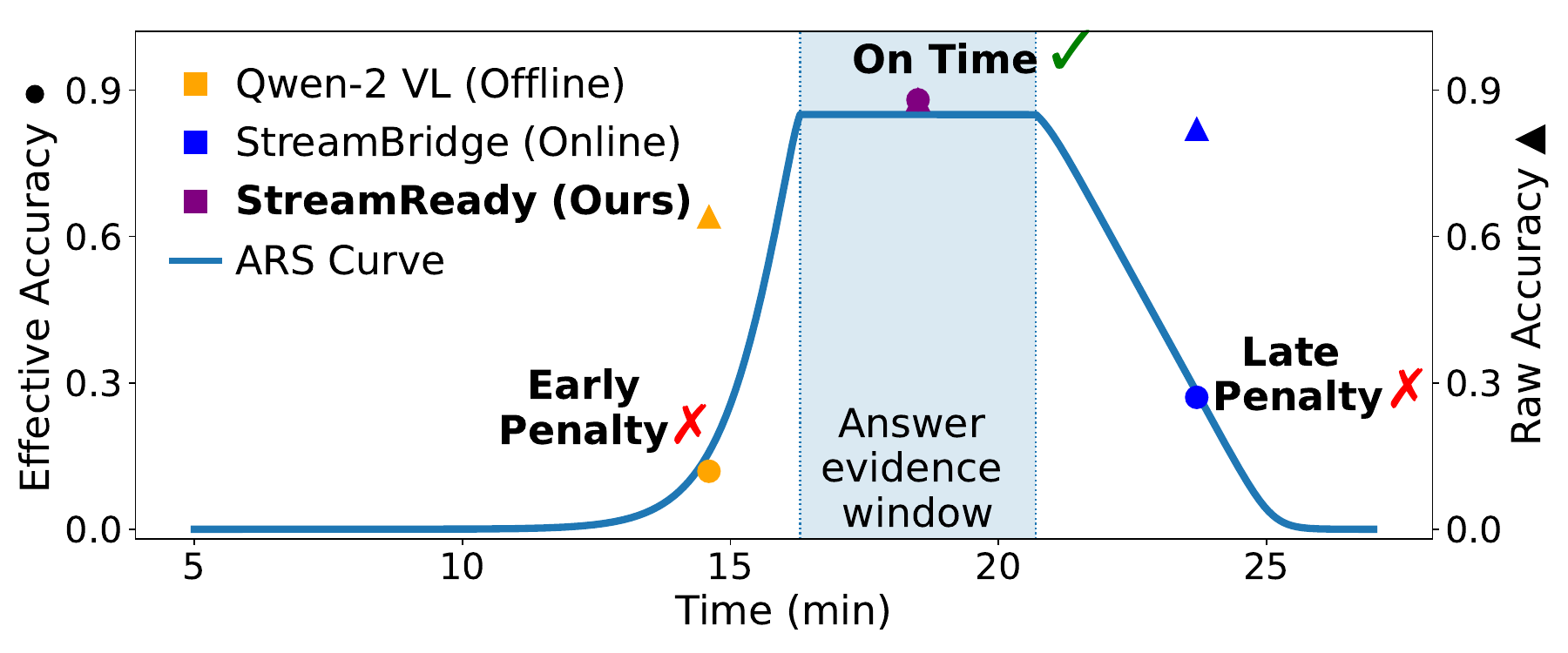}
    \captionof{figure}{\textbf{Readiness-aware streaming video understanding.} \textit{Left:} In proactive streaming settings, questions can precede their supporting evidence, requiring the model to monitor the evolving video and answer once the relevant cues appear. \textit{Right:} Under our readiness-aware formulation, effective accuracy jointly reflects answer correctness and timing via the Answer Readiness Score (ARS). Although all models achieve similar raw accuracy on this example, ARS reveals sharp performance drops for early (hallucinatory) or late (delayed) answers. In contrast, StreamReady responds within the evidence window, preserving high effective accuracy by answering at the appropriate moment.
    }      
    \label{fig:teaser}
\end{center}%
}]

\vspace{-5pt}
\begin{abstract}
Streaming video understanding often involves time-sensitive scenarios where
models need to answer exactly when the supporting visual evidence appears: answering before the evidence reflects speculation, answering after it has passed reduces real-time utility. 
To capture this behavior, we introduce a readiness-aware formulation of streaming video understanding with the \textbf{Answer Readiness Score (ARS)}, a timing-aware objective with asymmetric early and late penalties. When combined with correctness, ARS defines an effective accuracy that measures not just whether a model is right, but whether it answers at the appropriate moment.
Building on this formulation, we introduce \textbf{StreamReady}, a framework to unify temporal reasoning with on-time answering through  a lightweight readiness mechanism that decides if sufficient evidence has been observed before responding.
To evaluate this capability, we further introduce \textbf{ProReady-QA}, a benchmark with annotated answer evidence windows 
and proactive multi-turn questions across local and global contexts. StreamReady achieves superior performance on ProReady-QA, and consistently outperforms prior methods across eight additional streaming and offline long-video benchmarks, demonstrating robust and broadly generalizable video understanding capability.
\end{abstract}
\vspace{-22pt}    
\vspace{-6pt}
\section{Introduction}
\label{sec:intro}

Multimodal Large Language Models (MLLMs) have significantly advanced video understanding across diverse domains
\cite{li2025videomamba, chen2023videollm, li2024llava, qwen2vl, abdin2024phi, alayrac2022flamingo, azad2025disenq, schiappa2024probing, schiappa2024robustness}, particularly on short clips. However, their performance drops on long videos due to difficulties of reasoning over extended temporal context \cite{zou2024seconds}.
While recent efforts extend MLLMs to longer videos \cite{he2024ma, song2024moviechat+, cheng2024enhancing, azad2025hierarq, wang2024videoagent, zhi2025videoagent2}, these models still operate offline with full-video access during inference. In contrast, streaming video understanding represents an online variant of long-video reasoning, where frames arrive sequentially, evidence may appear before or after a question, and the model must operate without ever seeing the full video. This online capability is essential for real-world settings such as surveillance, sports analytics, robotics, and assistive systems that demand timely and context-aware reasoning. 

Building on advances in long-video reasoning, recent studies have begun exploring streaming video understanding
\cite{wang2024videollamb, zhang2024flash, qian2025dispider, wang2025streambridge, di2025rekv, kim2025infinipot, yang2025streammem}. 
However, most existing works focus on 
past-dependent (causal) reasoning where evidence is already available at question time, making answer correctness the main objective; leaving timing largely unexplored. 
In contrast, Many real-world scenarios require future-dependent (proactive) reasoning, where the question appears \emph{before} the supporting evidence, requiring the model to watch the unfolding video to determine when enough information has appeared (Figure \ref{fig:teaser}, \textit{left}).
In such proactive settings, models must prioritize answer timing as much as correctness: responding too early, even if correct, indicates unsupported speculation; responding too late causes unwanted delay. Developing this ability to identify the right moment to respond based on supporting evidence is therefore essential for truly effective streaming understanding.

Recently models have begun exploring such proactive behavior by deferring responses through auxiliary MLLMs \cite{wang2025streambridge} or prompt-based cues \cite{zhang2025avila}, though at the cost of nondeterminism or added compute. To evaluate this behavior, benchmarks \cite{lin2024streamingbench, li2025ovobench, wang2025omnimmi, wang2025proactivevideoqacomprehensivebenchmarkevaluating} include proactive scenarios to
encourage models to wait until relevant information appears.
However, they lack annotated answer evidence durations, making it difficult to verify if responses are given at an appropriate time. Consequently, these developments offer only a partial view of timing behavior since models may wait, but they lack any criteria for determining whether their chosen answer time is supported by the actual evidence.

To address these limitations, we formalize readiness-aware streaming video understanding, where the goal is not only to produce the correct answer, but to do so precisely \emph{when} sufficient evidence appears. At its core is the \textbf{Answer Readiness Score (ARS)}, a timing-aware evaluation metric with asymmetric penalties: a harsher early penalty discourages unsupported guesses before evidence, and a milder late penalty tolerates slight delays after the evidence ends (Figure \ref{fig:teaser} \textit{right}). Together, these penalties yield an effective accuracy that captures both answer correctness \emph{and} timing.
Building on this formulation, we propose \textbf{StreamReady}, a framework that unifies temporal reasoning with explicit answer-timing. Instead of relying on heavy auxiliary models or prompt heuristics, StreamReady introduces a lightweight learnable readiness token within its reasoning module, allowing the model to assess from its internal memory when sufficient evidence has appeared.
A small readiness head monitors this token, prompting the model to answer only when appropriate, ensuring responses are both accurate and timely.

To evaluate readiness-aware understanding, we introduce \textbf{ProReady-QA}, a benchmark designed for proactive scenarios with annotated answer evidence durations and tasks covering both local and global temporal contexts, enabling systematic evaluation under ARS. Our evaluations show that StreamReady effectively bridges the gap between raw and effective accuracy, outperforming existing methods in readiness-aware streaming settings.
Beyond ProReady-QA, StreamReady also achieves superior performance on other streaming benchmarks across proactive and non-proactive tasks, and generalizes well to offline long-video benchmarks. Together, our proposed formulation, method, and benchmark provide a unified foundation for advancing answer timing in streaming video understanding.

Our main contributions are as follows:
\begin{itemize}
\item We formalize readiness-aware streaming understanding and introduce the \textbf{Answer Readiness Score (ARS)} to jointly evaluate answer correctness and timing through asymmetric early and late penalties.
\item We propose \textbf{StreamReady}, a readiness-aware framework to integrate temporal reasoning with a readiness mechanism to decide evidence sufficiency before responding.
\item We develop \textbf{ProReady-QA}, a benchmark with annotated answer evidence windows and proactive multi-turn questions for evaluating timing behavior of streaming models.
\end{itemize}

\vspace{-6pt}
\section{Related Works}
\label{sec:rel_works}

\textbf{Offline Long Video Understanding with MLLMs.} Long-video understanding aims to model extended temporal context in videos spanning from few minutes to hours. Prior works use memory-based \cite{he2024ma, song2024moviechat+, cheng2024enhancing, balazevic2024memory} or agent-based approaches \cite{wang2024videoagent, zhi2025videoagent2, montes2025viqagent, kugo2025videomultiagents}, with some adopting query-conditioned storage \cite{wang2025videotree, azad2025hierarq}. However, these offline methods rebuild memory for every query and rely on full-video access, making them unsuitable for streaming settings where frames and questions arrive sequentially. Our work draws inspiration from query-aware conditioning but adapts it for streaming understanding without requiring memory reset.

\noindent \textbf{Streaming Video Understanding with MLLMs.} Inspired by long-video frameworks, memory-based \cite{qian2025dispider, zhang2024flashvstream, zeng2025streamforest, xiong2025streamchat, wang2025streambridge, wang2024videollamb, qian2024streaming} and retrieval-based approaches \cite{di2025rekv, yang2025streammem, kim2025infinipot} extend MLLMs to streaming by processing frames online and reusing past context. While effective for answer content, they lack mechanisms for deciding answer timing; particularly important in proactive scenarios where questions precede the answer. Our readiness-aware design complements these approaches by adding explicit answer timing, 
avoiding premature speculation and unnecessary delays by responding precisely when the evidence appears. 

\noindent
\textbf{Streaming Video Benchmarks.}
Existing streaming benchmarks \cite{xiong2025streamchat, zeng2025streamforest, huang2024videochatonline, li2024videochatflash, zhang2024flashvstream}
primarily support past-dependent question-answering, where timing has limited impact. More recent benchmarks include proactive scenarios \cite{lin2024streamingbench, wang2025proactivevideoqacomprehensivebenchmarkevaluating, li2025ovobench, wang2025omnimmi}, but remain limited to short clips and local context. In contrast, ProReady-QA supports proactive reasoning over long, continuous streams with both local and global multi-turn dependencies, enabling comprehensive evaluation of on-time answering in readiness-aware streaming.

\vspace{-6pt}
\section{Method}
\label{sec:method}
A fundamental requirement in streaming video understanding is not only \emph{what} a model answers but also \emph{when} it chooses to answer. We formalize this as \textbf{readiness-aware streaming video understanding}, where a model must produce the correct answer at the appropriate moment, supported by visual evidence. Unlike existing streaming setups that evaluate only correctness and therefore cannot distinguish on-time answers from mistimed ones, our formulation explicitly accounts for both answer content \emph{and} timing.

\subsection{Framework Overview}
\begin{figure}[t!]
    \centering
    \includegraphics[width=.92\linewidth]{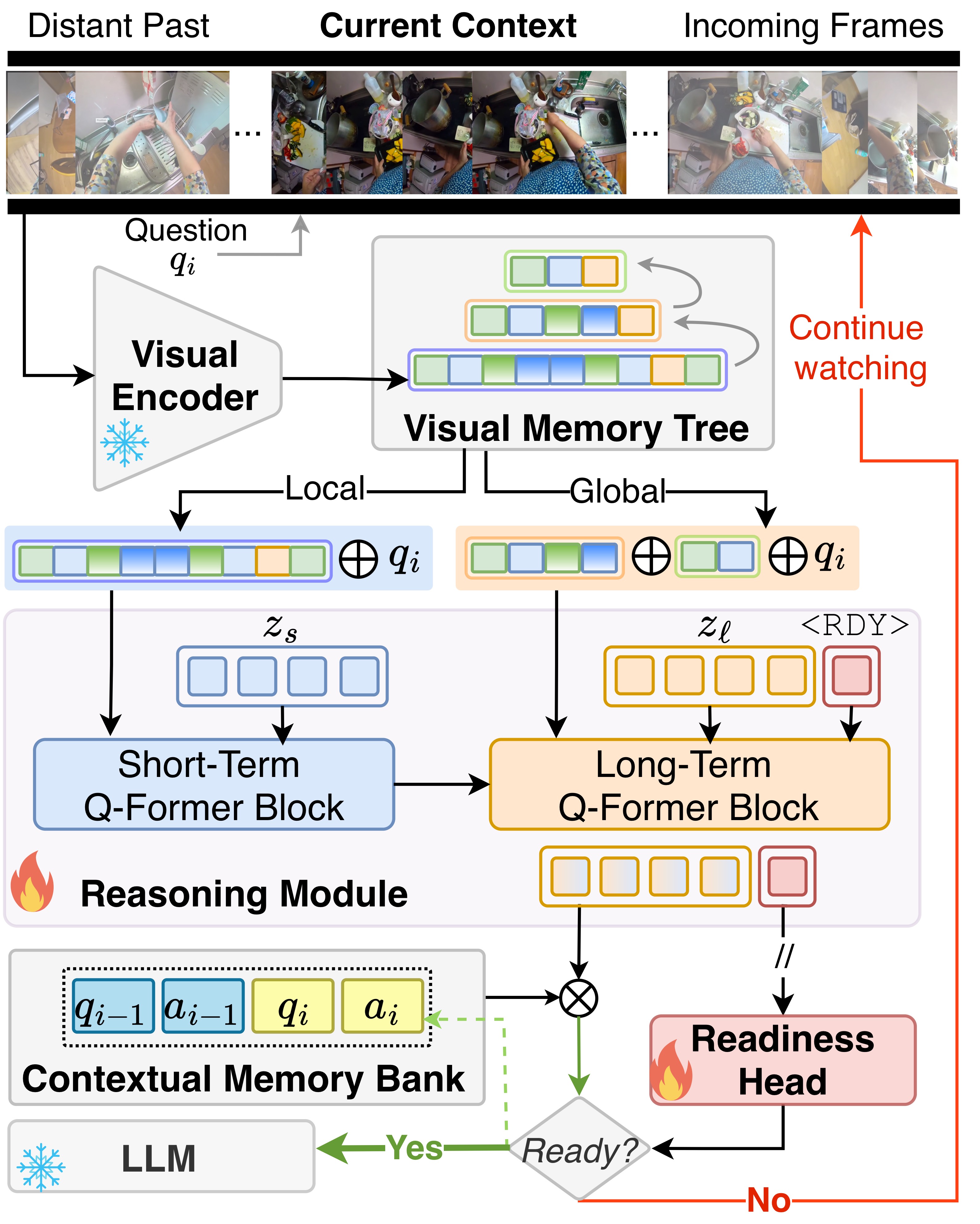}
    \captionsetup{aboveskip=1pt}
    \caption{\textbf{Framework Overview.} StreamReady encodes streaming videos into a visual memory tree and reasons through short and long-term branches. A learnable \texttt{<RDY>} token, guided by a readiness head,  
    gates the reasoning output until sufficient evidence is observed. 
    Once ready, the long-term representation, enriched with contextual information from past QA pairs, is sent to the LLM for answering, enabling readiness-aware streaming behavior.
    }
    \label{fig:overview}
    \vspace{-12pt}
\end{figure}

Building on this formulation, we introduce \textbf{StreamReady}, a readiness-aware framework that learns to determine the right moment to answer by monitoring the evolving video for supporting evidence. As the video unfolds, StreamReady stores them in a hierarchical memory (\S\ref{subsec:mem}), retrieves and reasons over temporally relevant context when a question appears (\S\ref{subsec:qf}) and uses a lightweight readiness mechanism (\S\ref{subsec:ans_ready}) to decide whether sufficient evidence is present to answer. If ready, it answers immediately; otherwise, it continues observing until the required evidence appears. Figure~\ref{fig:overview} provides an overview of the framework.

\subsection{Memory Storage}
\label{subsec:mem}
Streaming video understanding can benefit from both visual and semantic history, where visual cues guide perception and prior linguistic interactions provide context, enabling efficient reuse of past information. Since streaming videos contain fine-grained details and substantial temporal redundancy, an effective system must preserve key evidence while compacting redundant content.
StreamReady achieves this through two complementary memories: a Visual Memory Tree ($\mathcal{M}_{\text V}$) for multi-granular visual context, and a Contextual Memory Bank ($\mathcal{M}_{\text C}$) for long-range semantic dependencies across question–answer rounds.

\noindent \textbf{Visual Memory Tree ($\mathbf{\mathcal{M}_{\text{V}}}$).}
To efficiently represent long streaming videos, we maintain a multi-level  
Visual Memory Tree that progressively abstracts incoming frames. The lowest level $\mathcal{M}_{\text{V1}}$ stores the most recent frame embeddings 
in a FIFO buffer, preserving short-term details. Once full, its raw frames are compressed into a compact centroid set $\mathcal{M}_{\text{V2}} = \{c_1, c_2, \ldots, c_J\}$ via K-means clustering, forming a stable mid-level summary of the recent segment. As streaming continues, evicted frames $f_o$ of $\mathcal{M}_{\text{V1}}$, update $\mathcal{M}_{\text{V2}}$ through EMA-based clustering with decay factor $\alpha$: 
\begin{equation}
    c_j \leftarrow 
    \begin{cases}
        (1 - \alpha) c_j + \alpha f_{o},  &\text{if } \text{sim}(f_{o}, c_j) \geq \tau_t,\\
        \text{new centroid}, &\text{otherwise}
    \end{cases}
    \label{eqn:mv2}
    \vspace{-4pt}
\end{equation}
where, threshold $\tau_t$ tightens in stable scenes (favoring merges) and relaxes when novelty rises (allowing new clusters), keeping $\mathcal{M}_{\text{V2}}$ compact yet adaptive. 
When $\mathcal{M}_{\text{V2}}$ reaches capacity $J$ or shows distributional drift (e.g., frequent new-cluster creation), its centroids are abstracted into a coarse prototype set $\mathcal{M}_{\text{V3}} = \{s_1, s_2, \dots s_U\}$, also updated through EMA-based clustering to support fast continuous abstraction during stable scenes.
\begin{equation}
    \ s_u \leftarrow (1 - \alpha) s_u + \alpha \Big(\frac{1}{|\mathcal{I}_u|} \sum_{j \in \mathcal{I}_u}c_j \Big),
    \label{eqn:mv3}
    \vspace{-3pt}
\end{equation}
If centroids become heterogeneous (e.g., low mutual similarity or persistent novelty), a lightweight mini–K-means is triggered to realign prototypes, improving coherence without full re-clustering. 
This hierarchical design yields a compact multi-granular memory that preserves long-range fine-grained details while reducing redundancy, enabling efficient retrieval for query-aware reasoning.

\noindent \textbf{Contextual Memory Bank ($\mathbf{\mathcal{M}_{\text{C}}}$).} Beyond visual evidence, many streaming questions depend on earlier linguistic interactions. We support this with a Contextual Memory Bank $\mathbf{\mathcal{M}_\text{C}}$ where each entry stores the question embedding ($q_i$) and the learned representation ($a_i$) that was used to generate its answer, forming a lightweight semantic history, that
provides a complementary view to $\mathcal{M}_\text{V}$, enabling efficient context reuse for multi-turn reasoning.

\subsection{Query Aware Reasoning}
\label{subsec:qf}
Once a question $q_i$ arrives, the model transitions from passive encoding to active, query-aware retrieval and reasoning. Because evidence in streaming video is distributed unevenly over time, the model must reason over both recent and older fine-grained context and abstracted summaries. To support this, we use a dual-branch Q-Former \cite{azad2025hierarq}, consisting of a short-term branch $\mathcal{Q}_{\text{s}} (\cdot)$ that operates on $\mathcal{M}_{\text{V1}}$, and a long-term branch $\mathcal{Q}_{{\ell}} (\cdot)$, that operates on $\mathcal{M}_{\text{V2}}, \mathcal{M}_{\text{V3}}$.

\noindent \textbf{Short-term Reasoning.} To capture short-range query-relevant cues, the short-term branch $\mathcal{Q}_{\text{s}} (\cdot)$, attends to the raw frames in $\mathcal{M}_{\text{V1}}$, together with $q_i$, producing the short-term learned representation $z_s$: 
\begin{equation}
    z_s = \mathcal{Q}_{\text{s}}\big(\texttt{Concat}[\mathcal{M}_{\text{V1}}, q_i]\big), 
    \label{eqn:st}
\end{equation}

\noindent \textbf{Long-term Reasoning.} In long streaming videos, relevant evidence is often buried in earlier moments, but only a small portion might be relevant to any given query. To efficiently reason over useful long-range evidence, we perform coarse-to-fine query-aware retrieval over the higher levels of $\mathcal{M}_\text{V}$.

We first score all prototypes in $\mathcal{M}_{\text{V3}}$ and select the top-K high-level regions most likely to contain relevant evidence:
\begin{equation}
\mathcal{M}_{\text{V3}}^t
= \text{Top-K}\big((W_s q_i)^\top \mathcal{M}_{\text{V3}}\big),
\label{eq:topk}
\vspace{-4pt}
\end{equation}
This acts as a semantic map lookup that highlights the most relevant video regions. The prototype scores are normalized (via softmax) to stabilize routing and sharpen focus on plausible evidence.
For each selected prototype $s$, we then gather its associated centroids $c$ from $\mathcal{M}_{\text{V2}}$ to form a candidate pool $C_0$, refine their relevance score, and select the top-$m$ fine-grained slots:

\begin{equation}
\mathcal{M}_{\text{V2}}^t \;=\; \underset{c \in C_0}{\text{Top-m}}\; \big((W_c q_i)^\top c\big); \quad \ C_0 = \hspace{-5pt} \bigcup_{s \in \mathcal{M}_{\text{V3}}^t} \hspace{-5pt} \mathcal{M}_{\text{V2}}^{(s)},
\label{eq:topm}
\vspace{-6pt}
\end{equation}

Unlike the prototype selection stage, we avoid normalizing centroid scores since sharp ranking is essential for isolating specific evidence rather than broad regions. 
The retrieved prototypes $\mathcal{M}_{\text{V3}}^t$, centroids $\mathcal{M}_{\text{V2}}^t$, and the short-term learned representation $z_s$ are then passed to the long-term branch $\mathcal{Q}_{\ell}(\cdot)$ for reasoning:
\begin{equation}
    z_\ell = \mathcal{Q}_{\ell}\big(\texttt{Concat}[\mathcal{M}_{\text{V3}}^t, \mathcal{M}_{\text{V2}}^t, q_i], z_s \big) ,
    \label{eq:lt}
    \vspace{-6pt}
\end{equation}
This two-stage retrieval mirrors episodic recall: prototypes provide coarse temporal anchors, and centroids supply the fine-grained details.
Combining them with $z_s$ enables efficient long-term reasoning that is grounded in both broad and recent evidence, producing a focused representation.

\noindent \textbf{Contextual Reasoning.}
To effectively reuse past semantic context, we perform a contextual reasoning step over the contextual memory $\mathcal{M}_C$. The current question embedding is matched with stored previous question embeddings $\mathcal{M}_C$ to identify similar past QA interactions. A soft-gating mechanism selects the most relevant entries, and their answer representations are fused into the long-term visual feature $z_\ell$ through a lightweight cross-attention layer. This complements visual reasoning by incorporating prior semantic knowledge for coherent multi-turn understanding.

\subsection{Readiness Mechanism}
\label{subsec:ans_ready}
The reasoning modules determine \emph{what} to answer, but, readiness-aware streaming understanding also requires deciding \emph{when} to answer. To support this, we introduce a learnable \texttt{<RDY>} token and a Readiness Head. 

\noindent \textbf{Monitoring Readiness.}
We append the \texttt{<RDY>} token to the long-term reasoning representation $z_\ell$ within $\mathcal{Q}_{\ell}(\cdot)$ which already learns evidence from the visual memory tree, exposing the \texttt{<RDY>} token to the same evolving evidence used for answer reasoning. This ensures readiness decisions are grounded in the same representations that drive answer generation.
Before the relevant evidence appears, retrieval remains weakly aligned with the question, producing a diffused, low-confidence reasoning state. As supporting evidence arrives and retrieval becomes sharper and more question-consistent; the representation shifts towards an answer-bearing state. By residing within $\mathcal{Q}_{\ell}(\cdot)$, the \texttt{<RDY>} token naturally learns to track this transition and encode how prepared the model is to answer at a given time.

A lightweight Readiness Head, monitors this token and outputs a readiness score $R_{\text{pred}}\in[0,1]$ at each timestep. At inference, the model triggers the LLM to respond only when this score exceeds a threshold; otherwise, it continues observing future frames.
When the model is ready to answer, the fused representation of $z_\ell$ and $\mathcal{M}_C$ is passed to the LLM for response generation, and this fused representation is stored in $\mathcal{M}_C$ as the answer representation $a_i$ for the current question $q_i$. This gating enforces readiness-aware behavior by preventing premature guesses and unnecessary delays, ensuring responses occur precisely when the model judges the evidence to be sufficient.

\begin{figure*}[t!]
\centering
\begin{minipage}{0.48\linewidth}
    \centering
    \includegraphics[width=\linewidth]{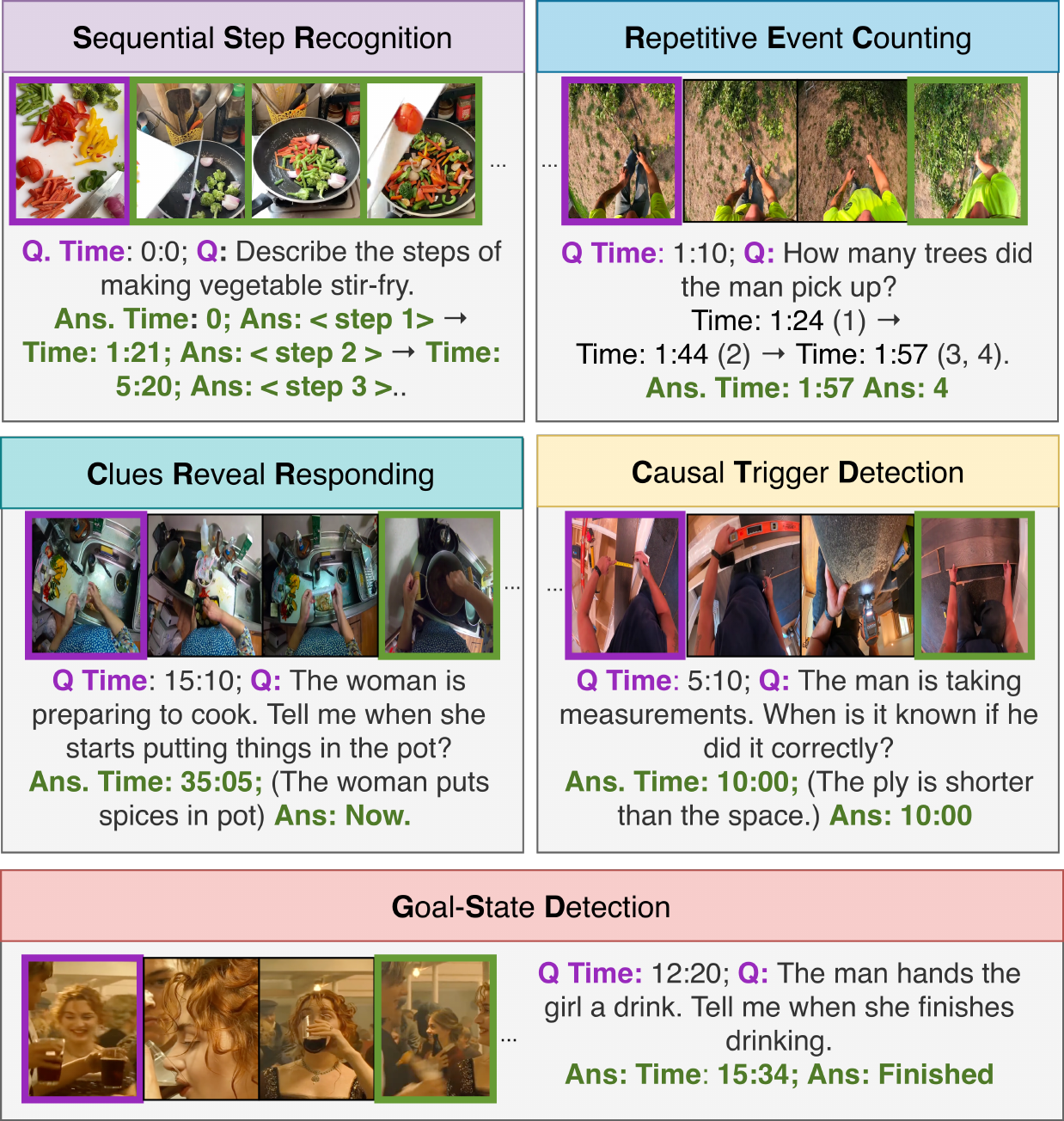}
    \caption{\textbf{Examples of each task in ProReady-QA.} Here, the \textcolor{purple}{\textbf{question}} and \textcolor{light_green}{\textbf{answer}} frames are color-coded.}
    \label{fig:benchmark}
\end{minipage}
\hspace{5pt}
\hfill
\begin{minipage}{0.5\linewidth}
    \centering
    \includegraphics[width=\linewidth, height=3.95cm]{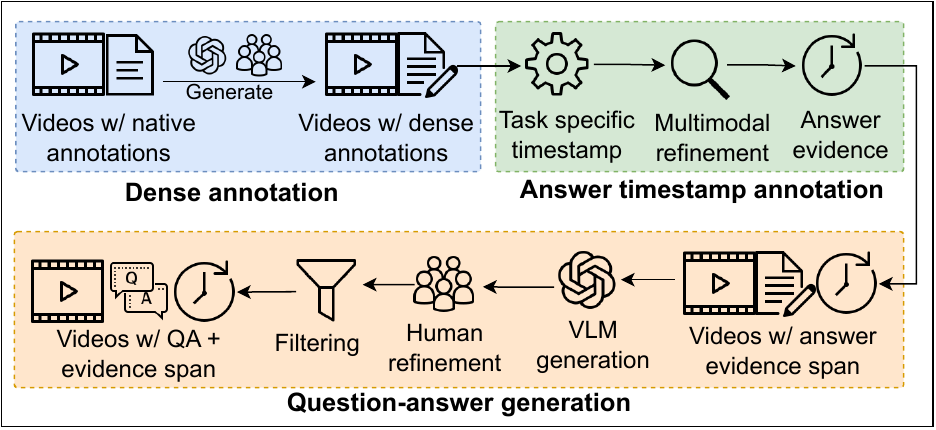}
    \caption{\textbf{Generation pipeline of ProReady-QA.} 
    \label{fig:benchmark_generation}
    }
    \vspace{-5pt} 
    \begin{table}[H]
    \centering
    \caption{\textbf{Comparison of ProReady-QA with prior benchmarks.}}
    \vspace{-5pt}
    \resizebox{\linewidth}{!}{
    \begin{tabular}{l|ccc ccc c}
    \toprule
         \multirow{2}{*}{\textbf{Benchmark}} & \textbf{QA} & \textbf{V Len.}  & \textbf{ V Avg.}&\multicolumn{2}{c}{\textbf{Multi-turn QA}} & \textbf{Ans.} \\
         & \textbf{(K)} & \textbf{(min)}& \textbf{(min)} &    \textbf{Local} & \textbf{Global} & \textbf{Evidence}\\
         \midrule
         \multicolumn{5}{l}{\textcolor{gray}{Non-Proactive Streaming Reasoning Benchmarks}}\\
         ODVBench \cite{zeng2025streamforest} &6.3 & $<$1-3 & 1.2 &\xmark & \xmark & \xmark \\
         StreamBench \cite{xiong2025streamchat}& 1.8 & 2-9 & 4.5&\xmark & \xmark & \xmark\\
         OVBench \cite{huang2024videochatonline}&7.0&$<$1-3 &1.5 & \xmark & \xmark & \xmark\\
         VStream-QA \cite{zhang2024flashvstream} &3.5& 30-60 & 40 & \xmark & \xmark & \xmark\\
         \midrule
         \multicolumn{5}{l}{\textcolor{gray}{Proactive Streaming Reasoning Benchmarks}}\\
         ProactiveVideoQA \cite{wang2025proactivevideoqacomprehensivebenchmarkevaluating} &1.4&$<$1-2 &2.1  & \xmark & \xmark & \xmark\\
         OVOBench \cite{li2025ovobench}&2.8&$<$1-30 & 7.1&\xmark & \xmark & \xmark\\
         Omni-MMI \cite{wang2025omnimmi} &2.3&$<$1-12 &5.4& \cmark & \xmark & \xmark\\
         StreamingBench \cite{lin2024streamingbench} &4.5 &1-10 &4.1 & \cmark & \xmark &\xmark\\
         \midrule
         \rowcolor{green!10}
         \textbf{ProReady-QA} &\textbf{5.0}&\textbf{30-60}&\textbf{40}  & \textbf{\cmark} & \textbf{\cmark} &\textbf{\cmark}\\
         \bottomrule
    \end{tabular}}
    \label{tab:benchmark_comparison}
\end{table}

\end{minipage}
\vspace{-13pt}
\end{figure*}

\noindent \textbf{Learning the Readiness Signal.}
Since the model must function online during inference but training has full-video access, we leverage this to construct weak pseudo-supervision without ever requiring ground-truth evidence timestamps. We estimate likely evidence locations by measuring similarity between the learned representation $z_\ell$ and centroid-level memory $\mathcal{M}_{\text{V2}}$ across time. $\mathcal{M}_{\text{V2}}$ offers a suitable balance between detail and compactness because its centroids capture fine-grained cues without the redundancy of $\mathcal{M}_{\text{V1}}$ or heavy abstraction of $\mathcal{M}_{\text{V3}}$.
High-similarity centroids define a pseudo-positive temporal region $P$, while low-similarity ones define a pseudo-negative region $N$.
The readiness mechanism is trained to assign higher readiness to $P$ than to $N$ through a pairwise contrastive loss:
\vspace{-4pt}
\begin{equation}
    \mathcal{L}_{ctr} = -log \ \sigma\ \big(R_{pred}(t^+) - R_{pred}(t^-)\big) , 
    \label{eq:lcont}
    \vspace{-4pt}
\end{equation}
where $t^+ \in P, t^- \in N$. Because $\mathcal{L}_{ctr}$ alone can produce noisy, unstable readiness signals, we add a mild temporal coherence regularizer in the final objective:
\begin{equation}
    \mathcal{L}_{rdy} = \mathcal{L}_{ctr} + \lambda_{reg}||\nabla_t R_{\text{pred}}(t)||_1 ,
    \label{eq:totloss}
    \vspace{-4pt}
\end{equation}
where $||\cdot||_1$ denotes the L1 norm. Crucially, $\mathcal{L}_{rdy}$ updates only the Readiness Head and the \texttt{<RDY>} token; gradients are stopped from the rest of the reasoning module so that the reasoning pathways learn \emph{what} to answer through its standard video-text loss, while the readiness mechanism independently learns \emph{when} to answer through the timing loss.

\section{ProReady-QA Benchmark and Evaluation}
\label{sec:benchmark}

To support readiness-aware streaming understanding evaluation, we introduce \textbf{ProReady-QA}, a benchmark designed specifically for long-duration streaming videos with proactive multi-turn questions and explicitly annotated answer evidence windows across local and global context (Table \ref{tab:benchmark_comparison}). ProReady-QA enables the first systematic evaluation of both answer correctness and temporal appropriateness, providing a dedicated testbed for studying answer timing behavior in streaming video models.

\subsection{Task Definition} 
\label{subsec:bench_task}
ProReady-QA spans five proactive reasoning tasks (Figure \ref{fig:benchmark}) requiring models to track future evidence and make temporally aware decisions:
(1) \textbf{Sequential Steps Recognition (SSR):} detect process transitions;
(2) \textbf{Repetitive Event Count (REC):} count recurring events;
(2) \textbf{Clues Reveal Responding (CRR):} answer once certain evidence appears;
(4) \textbf{Causal Trigger Detection (CTD):} detect cause–effect events;
(5) \textbf{Goal-State Detection (GSD):} identify goal completion.
The first three build on prior proactive tasks \cite{li2025ovobench, lin2024streamingbench}, while GSD and CTD extend proactive reasoning to higher-level temporal and causal understanding. 

\subsection{Dataset Construction}
ProReady-QA contains 10 one-hour Ego-4D \cite{grauman2022ego4d} and 22 half-hour MovieNet \cite{huang2020movienet} videos, with 5k proactive QA pairs with annotated answer evidence windows timestamps.

\noindent \textbf{Sourcing Videos.} We build upon long videos from VStream-QA \cite{zhang2024flashvstream}, which offer diverse activities and sufficient temporal depth to support future-dependent reasoning. However, VStream-QA’s questions are entirely past-dependent, making them ill-suited for proactive scenario.

\noindent \textbf{Generating QA and Evidence Timestamp.} We remove all past-dependent questions from VStream-QA and create new multi-turn proactive questions whose answers rely on {future} frames, following a semi-automatic QA generation pipeline \cite{zhang2024flashvstream, li2025ovobench} with human refinement. Complex tasks (CRR and CTD) include manually authored questions to capture nuanced temporal structure. Each QA pair is annotated with precise evidence windows by aligning native annotations with multimodal cues (visual frames, subtitles, action boundaries) to identify first- and last-valid evidence timestamps. For tasks with extended events (SSR, REC), we define onset and end times to cover the full target evidence duration, yielding fine-grained temporal boundaries essential for timing evaluation. ProReady-QA also incorporates local and global multi-turn dependencies, where later questions reference earlier entities or events, reflecting real-world long-horizon reasoning. Figure~\ref{fig:benchmark_generation} illustrates the overall QA generation process.

\begin{table*}[t!]
    \centering
    \caption{\textbf{Performance comparison of readiness-aware streaming understanding on ProReady-QA.} \dag: Qwen-2-VL backbone. Metrics include accuracy (Acc.), Answer Readiness Score (ARS), effective accuracy (Acc$_e$). \textbf{Best} and \underline{second-best} performances are highlighted. 
    }
    \label{tab:comparison_prored}
    \renewcommand{\arraystretch}{1.1}
    \resizebox{\linewidth}{!}{
    \begin{tabular}{l c| cc|cc|cc |cc|cc|ccc}
        \toprule
        \multirow{2}{*}{\textbf{Method}} & \multirow{2}{*}{\textbf{Size}} 
        &\multicolumn{2}{c|}{\textbf{SSR}}& \multicolumn{2}{c|}{\textbf{CRR}}& \multicolumn{2}{c|}{\textbf{REC}}& \multicolumn{2}{c|}{\textbf{GSD}}&\multicolumn{2}{c|}{\textbf{CTD}}& \multicolumn{3}{c}{\textbf{Average}}\\    
        && Acc. & ARS&Acc. & ARS&Acc. & ARS&Acc. & ARS&Acc. & ARS&Acc. & ARS & Acc$_e$ \\

        \midrule
        \multicolumn{14}{l}{\textcolor{gray}{\textit{Offline Video MLLMs}}} \\
        InternVL2 \cite{chen2024internvl2} & 8B &57.2&0.17&51.7&0.39&24.3&0.43&29.4&0.39&31.8&0.49&38.9&0.37&0.20 \\
        LLaVA-OneVision \cite{li2024llavaov} & 7B &71.8&0.48&58.2&0.53&22.3&0.45&39.4&0.24&34.8&0.18&45.3&0.38&0.29\\
        Qwen-2-VL \cite{qwen2vl}& 7B &67.9&0.32&53.1&0.39&20.7&0.31&35.1&0.52&30.3&0.28&41.4&0.34&0.20  \\
        LLaVA-NeXT-Video \cite{zhang2024llavanextvideo} & 7B &67.4&0.47&58.3&0.41&24.3&0.42&40.3&0.31&21.8&0.38&42.4&0.40&0.31 \\
    
        MiniCPM-V 2.6 \cite{yao2024minicpmv} & 8B &62.7&0.56&54.2&0.62&24.5&0.35&32.7&0.43&23.4&0.35&39.5&0.46&0.23 \\
        HierarQ \cite{azad2025hierarq} & 7B &67.4&0.32&55.8&0.45&28.7&0.44&45.2&0.38&32.8&0.43&46.0&0.40&0.27 \\
        \midrule
        
        \multicolumn{14}{l}{\textcolor{gray}{\textit{Online Video MLLMs}}} \\
        VideoLLM-online \cite{chen2024videollmonline} & 8B &53.2&0.30&51.1&0.38&18.3&0.12&30.5&0.34&24.9&0.44&29.6&0.32&0.18 \\
        Flash-VStream \cite{zhang2024flashvstream} \dag & 7B &71.4&0.61&60.3&0.59&22.4&0.31&42.8&0.48&21.4&0.35&43.7&0.47&0.34 \\
        Dispider \cite{qian2025dispider} \dag & 7B &47.2&0.59&52.3&0.55&20.3&0.20&34.2&0.33&27.2&0.41&36.2&0.42 &0.27 \\
        StreamForest \cite{zeng2025streamforest} \dag & 7B &65.8&0.41&57.4&0.51&24.4&0.24& 49.2&0.29&32.3&0.40&45.8&0.37&0.27   \\
        StreamBridge \cite{wang2025streambridge} \dag & 7B &\underline{72.2}&\underline{0.72}&59.7&\underline{0.65}&31.9&0.43&60.3&\underline{0.57}&41.4&\underline{0.49}&\underline{53.1}&\underline{0.60} & \underline{0.42}   \\
        ViSpeak  \cite{fu2025vispeak} \dag & 7B &67.5&0.61&55.2&0.54&25.3&0.27&48.2&0.49&27.3&0.23&44.7&0.43&0.31\\
        InfiniPot-V \cite{kim2025infinipot} \dag & 7B &69.4&0.50&\underline{60.8}&0.52&\underline{34.2}&\underline{0.54}&58.3&0.42&37.2&0.35&52.0&0.47&0.36 \\ 
        \midrule
        \rowcolor{green!10}
        \textbf{StreamReady \dag} & 7B &\textbf{74.3}&\textbf{0.78}&\textbf{63.3}&\textbf{0.73}&\textbf{39.6} & \textbf{0.68} & \textbf{61.2} & \textbf{0.68} & \textbf{43.5} & \textbf{0.59} & \textbf{56.4} & \textbf{0.69}&\textbf{0.53}\\

        \bottomrule
    \end{tabular}
    }
\end{table*}

\subsection{Evaluating Answer Readiness}
\label{subsec:ep_lp_description}
To evaluate readiness-aware streaming performance, we introduce the \textbf{Answer Readiness Score (ARS)}, a timing-aware evaluation metric that penalizes answers given too early or too late relative to their evidence. When combined with accuracy (Acc), ARS produces an effective accuracy (Acc$_e$) that rewards predictions that are both correct and well-timed. For a set of $N$ questions, ARS is defined as: 
\begin{equation}
\text{ARS} = \frac{1}{N} \sum_{i=1}^N (\text{EP}_i \cdot \text{LP}_i); \quad \text{Acc}_e = \text{Acc} \times  \text{ARS}
\vspace{-3pt}
\end{equation}
where $\text{EP}$ and $\text{LP}$ represent early and late penalties. 

For each question, ProReady-QA provides a ground-truth evidence window $[t_s, t_e]$, where $t_s$ marks when sufficient evidence first appears and $t_e$ when it ceases to be valid. Given a model answers at time $t_a$, these intervals enable ARS to evaluate timing behavior through complementary early and late penalties aligned with our readiness-aware formulation.

\noindent\textbf{Early Penalty (EP).}
Answering \emph{before} any supporting evidence appears is the most severe readiness failure, as it reflects speculation rather than observation. To discourage such behavior, the Early Penalty sharply decreases as the model answers earlier than the evidence onset $t_s$, scaled by the median evidence duration $\tau$ for consistency.
\begin{equation}
    \text{EP} = \text{softmin}\left(1, 2 \ \sigma \left( \gamma_e \dfrac{t_a-t_s}{\tau+ \epsilon} \right) \right)
    \label{eq:ep}
\end{equation}
Here, $\sigma(\cdot)$ is sigmoid, $\gamma_e$ controls penalty sharpness, $\epsilon$ ensures numerical stability. $\text{EP} \to 1$ as $t_a$ approaches $t_s$, while early answers yield lower scores.

\noindent\textbf{Late Penalty (LP).}
After the evidence ends at $t_e$, delayed responses indicate hesitation rather than hallucination. To capture this milder form of readiness error, the Late Penalty gently decreases with delay, encouraging timely responses without overly penalizing slight delays:
\begin{equation}
    \text{LP} = \text{softmin}\left(1, \text{softmax}\left( 0, 1-\gamma_\ell\dfrac{t_a-t_e}{\tau+ \epsilon} \right) \right)
    \label{eq:lp}
\end{equation}
where $\gamma_\ell$ controls decay slope, keeping $\text{LP}$ near 1 for minor delays and lower for prolonged ones.
For questions with multiple valid answers, we compute ARS separately for each turn and report their average. 

\vspace{-6pt}
\section{Experiments}
\label{sec:exp}
\begin{table*}[t!]
    \centering
    \caption{\textbf{Performance comparison of streaming video understanding.} \dag: Qwen-2-VL backbone. Accuracy is the reported metric.}
    \vspace{-5pt}
    \begin{minipage}{.67\textwidth}
    \subcaption{Benchmark: StreamingBench (Strm), and OVOBench (OVO).}
    \label{tab:comparison_no_ovb}
    \renewcommand{\arraystretch}{1.13}
    \resizebox{\linewidth}{!}{
    \begin{tabular}{l c| cc|cc cc|cc}
        \toprule
        \multirow{3}{*}{\textbf{Method}} & \multirow{3}{*}{\textbf{Size}} & \multicolumn{2}{c|}{\textbf{Proactive:  \cmark }}& \multicolumn{4}{c|}{\textbf{Proactive:  \xmark }} &\multicolumn{2}{c}{\textbf{Average}}\\
        \cline{3-10}
        &&\textbf{Strm} & \textbf{OVO}& \multicolumn{2}{c}{\textbf{Strm}} & \multicolumn{2}{c|}{\textbf{OVO}} &  \multirow{2}{*}{\textbf{Strm}} & \multirow{2}{*}{\textbf{OVO}}  \\
        &  & Cont.&Fwd.&Real & Omni & Real & Back  \\
        \midrule
        \multicolumn{7}{l}{\textcolor{gray}{\textit{Offline Video MLLMs}}} \\
        InternVL2 \cite{chen2024internvl2} & 8B &32.4&45.4&63.7 &35.8& 60.7 & 44.0 &44.0&50.0  \\
        LLaVA-OneVision \cite{li2024llavaov} & 7B &32.7&50.9& 71.1 &38.4&62.8&45.0 & 47.4&52.9   \\
        Qwen-2-VL \cite{qwen2vl}& 7B &31.7&48.9& 69.0 &34.9&60.7&48.6 & 45.2& 52.7  \\
        LongVU \cite{shen2024longvu} & 7B & -&48.5&- &-&57.4&39.5 & - & 48.5   \\
        LongVA \cite{zhang2024longva} & 7B & 30.2&-&63.1&35.9&-&-&43.1&-\\
        LLaVA-NeXT-Video \cite{zhang2024llavanextvideo} & 7B &34.3&54.2& 69.8&41.7&63.3&41.7&48.6&53.1 \\
        VITA 1.5 \cite{fu2025vita} & 7B &27.4&53.5& 52.3 & 33.1 &63.5&41.5&37.6&52.8 \\
        MiniCPM-V 2.6 \cite{yao2024minicpmv} & 8B & 35.0 & - & 67.4 & 35.0 & - &- &45.8&-\\
        HierarQ \cite{azad2025hierarq} & 7B & 35.1& 48.3 & 69.7& 44.4 & 67.3 & 48.3&49.7&54.6\\
        \midrule
        \multicolumn{7}{l}{\textcolor{gray}{\textit{Online Video MLLMs}}} \\
        VideoLLM-online \cite{chen2024videollmonline} & 8B &26.6&-& 36.0 &28.5&20.8&17.7&30.4&19.3   \\
        Flash-VStream \cite{zhang2024flashvstream} \dag & 7B &24.1&44.2&23.2 &26.0&29.9&25.4&24.4&33.2 \\
        Dispider \cite{qian2025dispider} \dag & 7B &33.6&34.7& 67.6 &35.7&54.6&36.1 & 45.6&41.8 \\
        StreamForest \cite{zeng2025streamforest} \dag & 7B & - &53.5&\underline{77.3} &-&61.2&52.0 &-&55.6 \\
        ReKV \cite{di2025rekv} \textcolor{gray}{w/o offload}  & 7B &30.7&69.1 &37.4&-&-&- &-&-\\
        StreamBridge \cite{wang2025streambridge} \dag & 7B &32.6&48.4& 77.0 &24.1&\underline{71.3}&\underline{68.1} & 44.6&\underline{62.6}  \\
        ViSpeak  \cite{fu2025vispeak} \dag & 7B &\underline{43.9}&\underline{54.3}& 70.4 &\underline{61.6}&66.3&57.5 &\underline{58.6}&59.4 \\
        InfiniPot-V \cite{kim2025infinipot} \dag & 7B & - & 47.9&76.4&-&65.9&47.6&-&53.8\\ 
        TimeChat-Online \cite{yao2025timechat} & 7B & 35.3 & 36.4 & 75.4 & 37.8& 58.6 & 42.0&49.5&45.7\\
        StreamAgent \cite{yang2025streamagent} \dag & 7B & 34.6 & 45.4 & 74.3&36.3&61.3&41.7&48.4&49.5\\
        \midrule
        \rowcolor{green!10}
        \textbf{StreamReady \dag} & 7B & \textbf{48.2} & \textbf{58.8} & \textbf{78.3} & \textbf{63.7} & \textbf{73.6} & \textbf{72.2} & \textbf{63.4}&\textbf{68.2} \\
        \bottomrule
    \end{tabular}
    }
    \end{minipage}
    \hfill
    \begin{minipage}{.3\textwidth}
        \subcaption{Benchmark: VStream-QA. Non-Proactive. 
        }
    \label{tab:comparison_vstream}
    \renewcommand{\arraystretch}{.92}
    \resizebox{\linewidth}{!}{
    \begin{tabular}{lc|cc}
    \toprule
        \textbf{Method} & \textbf{Size} &\textbf{RE} & \textbf{RM}\\
        \midrule
        \multicolumn{3}{l}{\textcolor{gray}{\textit{Online Video MLLMs}}}\\
         MovieChat \cite{song2024moviechat} & 7B & 50.7 & 36.0\\
         HierarQ \cite{azad2025hierarq} & 7B & 56.4 & 49.4 \\
         Flash-VStream \cite{zhang2024flash} \dag & 7B & 57.3 & \underline{53.1}\\
         ReKV \cite{di2025rekv} \textcolor{gray}{w/o offload} & 7B & 55.8 & 50.8\\
         InfiniPot-V \cite{kim2025infinipot} \dag & 7B & \underline{57.9} & 51.4 \\
         \midrule
         \rowcolor{green!10}
         \textbf{StreamReady \dag} & 7B & \textbf{64.8} & \textbf{57.2}  \\
         \bottomrule
    \end{tabular}}
    \vspace{1mm}
    \subcaption{Benchmark: OVBench. Non-Proactive.}
    \label{tab:comparison_ovb_only}
    \renewcommand{\arraystretch}{.92}
    \resizebox{\linewidth}{!}{
    \begin{tabular}{l c|c}
        \toprule
        \textbf{Method} & \textbf{Size} & \textbf{Avg.} \\
        
        \midrule
        \multicolumn{3}{l}{\textcolor{gray}{\textit{Offline Video MLLMs}}} \\
        LLaMA-VID \cite{li2024llamavid} & 7B & 41.9 \\
        MiniCPM-V2.6 \cite{yao2024minicpmv} & 8B & 39.1\\
        LITA \cite{huang2024lita} & 7B & 20.4\\
        InternVL2 \cite{chen2024internvl2} & 8B & 48.7 \\
        LongVA \cite{zhang2024longva} & 7B & 43.6 \\
        LLaVA-OneVision \cite{li2024llavaov} & 7B & 49.5 \\
        Qwen-2-VL \cite{qwen2vl} & 7B & 48.7 \\
        TimeChat \cite{ren2024timechat} & 7B & 12.8\\
        HierarQ \cite{azad2025hierarq}& 7B& 57.8\\
        \midrule
        \multicolumn{3}{l}{\textcolor{gray}{\textit{Online Video MLLMs}}} \\
        VideoLLM-Online \cite{chen2024videollmonline} & 8B & 9.6\\
        MovieChat \cite{song2024moviechat} & 7B & 30.9 \\
        VideoChat-Online \cite{huang2024videochatonline} & 4B & 54.9 \\
        Flash-VStream \cite{zhang2024flashvstream} \dag & 7B & 31.2\\
        StreamForest \cite{zeng2025streamforest} \dag & 7B & \underline{60.5} \\
        \midrule
        \rowcolor{green!10}
        \textbf{StreamReady \dag} & 7B & \textbf{63.9} \\
        \bottomrule
    \end{tabular}
    }
    \end{minipage}
    \vspace{-10pt}
\end{table*}

\noindent \textbf{Datasets and Metrics.}
We evaluate StreamReady on ProReady-QA using accuracy, ARS and effective accuracy. To show generalization beyond readiness-aware streaming, we further evaluate on four streaming benchmarks: StreamingBench \cite{lin2024streamingbench}, OVOBench \cite{li2025ovobench}, OVBench \cite{huang2024videochatonline}, and VStream-QA \cite{zhang2024flashvstream}; and four offline long-video benchmarks: VideoMME \cite{fu2024videomme}, MLVU \cite{zhou2024mlvu}, MVBench \cite{li2024mvbench}, and EgoSchema \cite{mangalam2023egoschema}, following each dataset’s official evaluation protocol and reporting accuracy.

\noindent \textbf{Implementation Details.}
We use Qwen-2 VL \cite{qwen2vl} as the backbone model and initialize the dual-branch Q-Former using pretrained weights \cite{azad2025hierarq}. Offline models are tested on ProReady-QA with progressively truncated prefixes of the full video and recording when the correct answer first appears, while streaming models process frames sequentially with the prompt ``Answer whenever you are ready” following \cite{lin2024streamingbench, li2025ovobench}.  For offline long-video benchmarks, readiness and contextual reasoning are disabled. We set $\gamma_e=6$, $\gamma_\ell=1$ for ARS penalties and use a readiness threshold of 0.35 for LLM trigger following \cite{fu2025vispeak, wang2024videollm, wang2025streambridge}.

\begin{table}[t!]
    \centering
    \caption{\textbf{Performance comparison of offline long-video understanding.} \dag: Qwen-2-VL backbone. Accuracy is reported.}
    \label{tab:comparison_offline}
    \vspace{-3pt}
    \resizebox{\linewidth}{!}{
    \begin{tabular}{l c| c c c c}
        \toprule
        \textbf{Method} & \textbf{Size} &  \textbf{VidMME} & \textbf{MLVU} & \textbf{MVB} & \textbf{EgoSch} \\
        \midrule
        \multicolumn{6}{l}{\textcolor{gray}{\textit{Open-source Offline Video MLLMs}}} \\
        VideoChat-GPT \cite{maaz-etal-2024-video} & 7B &  - &  31.3 &32.7  & 49.6 \\
        LLaMA-VID \cite{li2024llamavid} & 7B & 33.2 & 41.9 & 38.5 & 38.5 \\
        InternVL2 \cite{chen2024internvl2} & 8B & 54.0 & 64.0 & 65.8 & 55.0 \\
        LongVA \cite{zhang2024longva} & 7B & 52.6 & 56.3 &51.3  & 46.7  \\
        LLaVA-OneVision \cite{li2024llavaov} & 7B & 58.2 & 64.7 & 56.7 & 60.1 \\
        Qwen-2-VL \cite{qwen2vl} & 7B & 63.3 & 65.8 & 67.0 & 66.7 \\
        LongVU \cite{shen2024longvu} & 7B & 60.6 & 65.4 & 66.9 & 67.6  \\
        LLaVA-Video \cite{zhang2024llavavideo} & 7B & 63.3 & \underline{70.8} & 58.6 & 57.3 \\
        HierarQ \cite{azad2025hierarq} & 7B & 63.7 & 69.4 & 67.6 & \underline{67.3}\\
        \midrule
        \multicolumn{6}{l}{\textcolor{gray}{\textit{Open-source Online Video MLLMs}}} \\
        MovieChat \cite{song2024moviechat} & 7B & 38.2 & 25.8  & 55.1 & 53.5 \\
        Flash-VStream \cite{zhang2024flashvstream} \dag & 7B & 61.2 & 66.3 & 65.4 & 68.2 \\
        VideoChat-online \cite{huang2024videochatonline} & 4B & 52.8 & 60.8 & 64.9 & 54.7 \\
        Dispider \cite{qian2025dispider} \dag & 7B & 57.2 & 61.7 & - & 55.6 \\
        StreamForest \cite{zeng2025streamforest} \dag & 7B & 61.4 & 70.0 & \underline{70.2} & - \\
        ReKV \cite{di2025rekv} \textcolor{gray}{w/o offload}  & 7B & - & 68.5 & - & 60.7 \\
        InfiniPot-V \cite{kim2025infinipot}  \dag & 7B & 62.8 & 65.8 & - & 65.6 \\
        VideoLLaMB \cite{wang2024videollamb} & 7B&41.4&-&52.5&-\\
        TimeChat-Online \cite{yao2025timechat} & 7B & 62.5 & 65.4 & - & -  \\
        StreamBridge \cite{wang2025streambridge}  \dag & 7B & \underline{64.4} & 69.6 & 64.4 & 66.9 \\
        ViSpeak \cite{fu2025vispeak} \dag & 7B & 55.0 & 54.1 & 54.1 & - \\
        \midrule
        \rowcolor{green!10}
        \textbf{StreamReady \dag \textcolor{gray}} & 7B & \textbf{65.8} & \textbf{71.3} & \textbf{71.8} & \textbf{70.4}\\
        \bottomrule  
    \end{tabular}
    }
    \vspace{-20pt}
    \end{table}

\begin{table*}[t!]
    \centering
    \begin{minipage}{.52\textwidth}
        \centering
        \captionsetup{aboveskip=2pt}
    \caption{\textbf{Ablation studies} for each component.}
    \renewcommand{\arraystretch}{.9}
    \resizebox{\linewidth}{!}{
    \begin{tabular}{l|cc|cc|cc}
    \toprule
         \multirow{2}{*}{Method} & \multicolumn{2}{c|}{\textbf{REC}} & \multicolumn{2}{c|}{\textbf{GSD}} & \multicolumn{2}{c}{\textbf{CTD}} \\
         & Acc. & ARS & Acc. & ARS & Acc. & ARS\\
         \midrule
         \multicolumn{7}{l}{\textcolor{gray}{Contribution of each component}}\\
         Baseline &20.7&0.31& 35.1 & 0.52 & 30.3 & 0.28 \\
         + Trivial Reasoning \cite{azad2025hierarq} &28.7&0.44&48.2 & 0.53 & 34.2 & 0.36\\
         $\qquad$ + Readiness Mechanism & 28.4 & 0.50 & 47.9 & 0.58 & 34.3 & 0.42 \\
         + Memory Storage &32.4&0.46&50.2 & 0.52 & 38.8 & 0.36  \\
         $\qquad$ + Query-aware Reasoning &39.4&0.48&60.9 & 0.53 & \textbf{43.6} & 0.39\\
        \rowcolor{green!10}
         \textbf{$\qquad$ + Readiness Mechanism} &\textbf{39.6}&\textbf{0.68}& \textbf{61.2} & \textbf{0.68} & \underline{43.5} & \textbf{0.59} \\
         \midrule
          \multicolumn{6}{l}{\textcolor{gray}{Design choice of Readiness Mechanism }}\\
          Only MLP & 39.5 & 0.52& 60.9 & 0.55 & 42.9 & 0.42 \\
          LLM w/ Heuristic Triggers \cite{zhang2025avila} & 39.6 & 0.54 & 61.4 & 0.54 & 43.5 & 0.43\\
          Auxilliary MLLM \cite{wang2025streambridge} & 39.2 & 0.60&\textbf{61.2}&0.61&43.3 & 0.46\\
          \texttt{<RDY>} + Head (Transformer) & 39.5 & \textbf{0.69} &\textbf{61.6} & \textbf{0.68} & 43.1 & {0.58 }\\
          \rowcolor{green!10}
          \textbf{\texttt{<RDY>}} \textbf{+ Head (MLP)} &\textbf{39.6}&\underline{0.68}& \underline{61.2} & \textbf{0.68} & \textbf{43.5} & \textbf{0.59}\\
         \midrule
         \multicolumn{6}{l}{\textcolor{gray}{Placement of \texttt{<RDY>} token}}\\
         Input of $\mathcal{Q}_\text{s}$ & 39.1 & 0.31&60.9 & 0.51 & 43.1 & 0.18 \\
         Learned representation of $\mathcal{Q}_\text{s}$ & 39.6 & 0.38 & \textbf{61.3} & 0.54 & 42.8 &0.24 \\
         Input of $\mathcal{Q}_\ell$ & 39.4 & 0.54& 61.1 &0.62 & \textbf{43.6}&0.49 \\
        \rowcolor{green!10}
          \textbf{Learned representation of $\mathcal{Q}_\ell$} &\textbf{39.6}&\textbf{0.68}& \underline{61.2} & \textbf{0.68} & \underline{43.5} & \textbf{0.59}\\
        \bottomrule
    \end{tabular}
    }
    \label{tab:ablation}
    \end{minipage}  
    \hspace{2pt}
    \begin{minipage}{.46\textwidth}
    \begin{minipage}{\textwidth}
        \includegraphics[width=\linewidth]{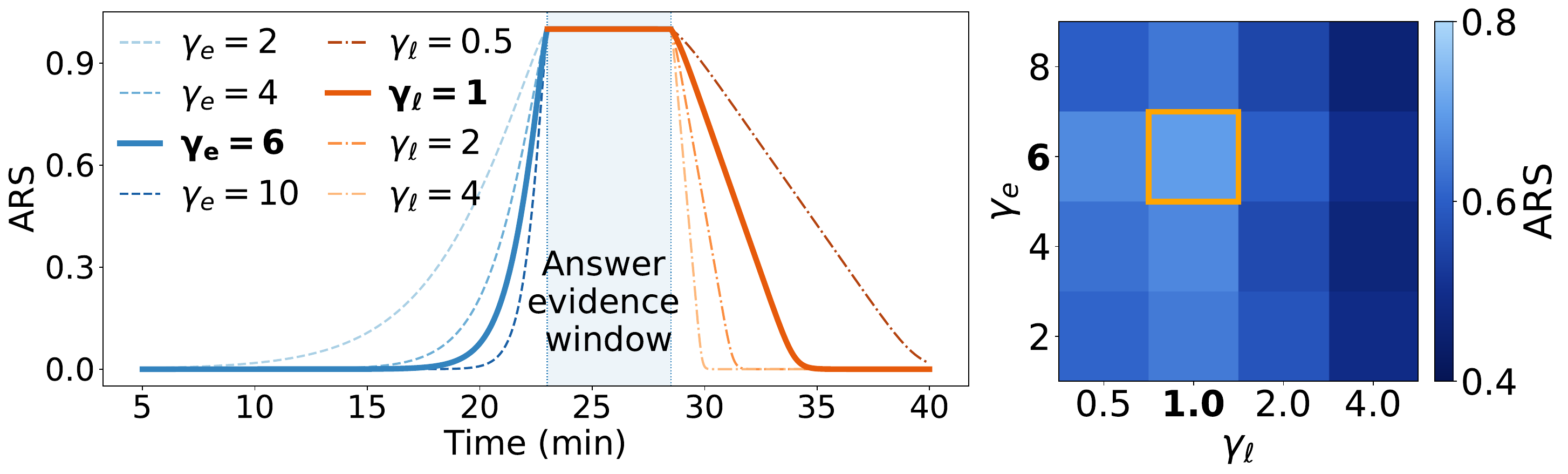}
    \captionsetup{aboveskip=2pt}
        \captionof{figure}{\small \textbf{Penalty sharpness for early and late responses.} \textit{Left:} Readiness curves for different penalty strengths. \textit{Right:} Resulting ARS, with selected $\gamma_e, \gamma_\ell$ combination highlighted.}
        \label{fig:ars_sharpness}
    \end{minipage}
    \vspace{2pt}
    \begin{minipage}{\textwidth}
        \includegraphics[width=\linewidth,]{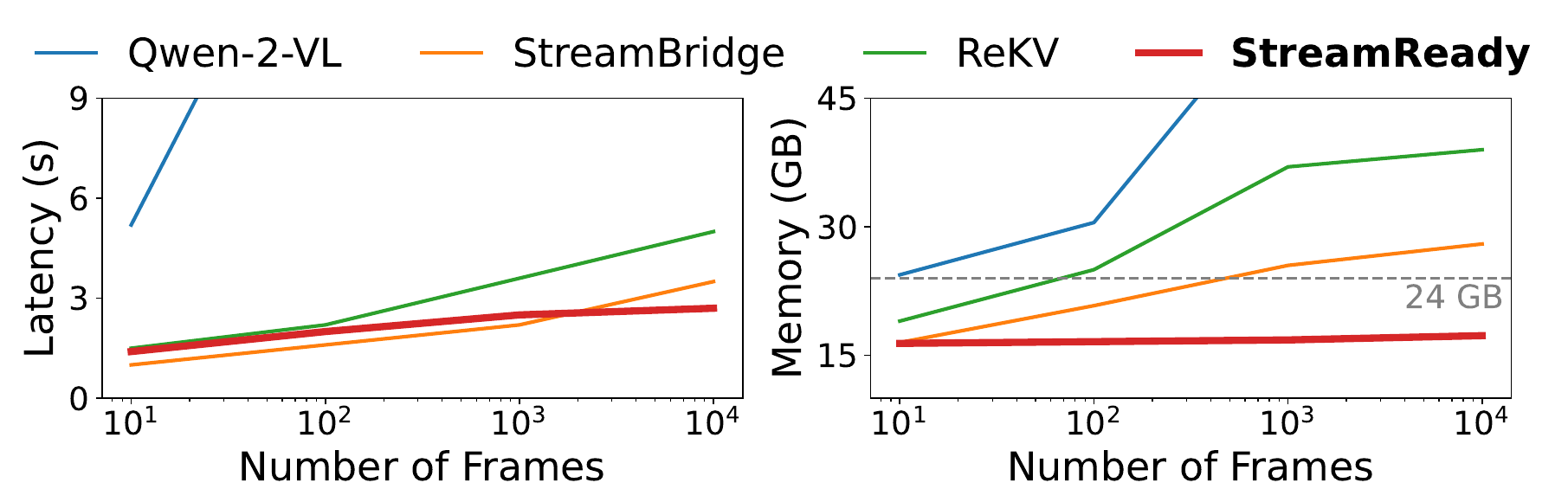}
    \captionsetup{aboveskip=2pt}
    \captionof{figure}{\textbf{Latency and memory usage analysis.} }
        \label{fig:compute}
    \end{minipage}
    \end{minipage}
    \vspace{-15pt}
\end{table*}

\subsection{Results}

\textbf{Readiness-Aware Streaming Understanding.} Table~\ref{tab:comparison_prored} shows that StreamReady achieves the highest accuracy and ARS across all five ProReady-QA tasks, surpassing the best model by \mytexttildenew3\% in accuracy and \mytexttildenew9\% in ARS on average, with the largest ARS gains on REC, GSD, and CTD tasks. StreamReady’s readiness mechanism  reduces mistimed responses, producing tighter temporal alignment that directly lifts ARS. Additionally, its hierarchical visual memory exposes the model to clearer evidence boundaries, making readiness estimation more reliable and leading to a smaller gap between raw and effective accuracy.

\vspace{-1pt}
\noindent \textbf{Streaming video understanding.} Tables~\ref{tab:comparison_no_ovb}–\ref{tab:comparison_ovb_only} shows StreamReady consistently outperforms prior models, achieving up to \mytexttildenew5\% gains on proactive tasks of streaming benchmarks. 
While the readiness mechanism improves proactive behavior by encouraging the model to wait for sufficient evidence, most of the accuracy improvements in these benchmarks come from StreamReady's stronger evidence retrieval and long-horizon reasoning. Its visual memory offers reliable access to relevant visual cues, and the contextual memory provides complementary semantic history for multi-turn interactions. Combined with query-aware long-range reasoning, this design suppresses irrelevant historical frames and emphasizes the segments most predictive of an answer, leading to more stable and accurate predictions even when timing itself is not evaluated.

\noindent \textbf{Offline long video understanding.} Table~\ref{tab:comparison_offline} shows that StreamReady also maintains strong performance in offline long-video understanding, outperforming prior models despite temporal awareness being irrelevant. Here, the benefit mainly comes from its ability to integrate long-range visual structure as the memory hierarchy offers compact yet expressive summaries of extended video segments, allowing the model to reason accurately over long temporal spans. 

\subsection{Ablation}
We ablate StreamReady's readiness mechanism on three challenging ProReady-QA tasks (REC, GSD, and CTD). Additional architectural ablations are in Supplementary.

\noindent \textbf{Contribution of each component.} Table~\ref{tab:ablation} (\textit{top}) shows that while adding basic reasoning \cite{azad2025hierarq} improves accuracy, it provides little timing benefit. Incorporating the readiness mechanism on it increases ARS modestly, but major gains occur with our stronger memory and reasoning modules. Together, these components enhance both accuracy and timing, showing that improved evidence retrieval strengthens not only \emph{what} but also \emph{when} the model answers.

\noindent \textbf{Design choice of readiness mechanism.} 
Table~\ref{tab:ablation} (\textit{middle}) shows that an MLP-based head is insufficient to judge readiness, while LLM or MLLM-based readiness adds only modest improvement due to weak coupling with reasoning. Embedding the \texttt{<RDY>} token within the reasoning module offers the best and most stable ARS improvements by directly accessing evolving evidence, with a lightweight MLP head matching Transformer performance at lower cost.

\noindent \textbf{Design choice of \texttt{<RDY>} token placement.}
Table~\ref{tab:ablation} (\textit{bottom}) shows that placing \texttt{<RDY>} in the short-term branch yields noisy timing signals due to its limited local context, while positioning it as input to the long-term branch offers moderate but unstable results as it only passively attends to retrieved evidence. The best ARS is achieved by attaching \texttt{<RDY>} to the learned long-term representation, where it co-evolves with query-aligned evidence, enabling the most reliable readiness detection.

\vspace{-5pt}
\subsection{Analysis}
\noindent \textbf{Penalty Sharpness for Early and Late Responses.} Figure \ref{fig:ars_sharpness} (\textit{left}) shows how the sharpness parameters ($\gamma_e$, $\gamma_\ell$) influence readiness behavior and ARS. Larger $\gamma_e$ effectively suppresses premature answers, while smaller values may over-reward early guesses. Conversely, lower $\gamma_\ell$ gently tolerates slight delays, whereas higher values penalize hesitation too strongly. Figure \ref{fig:ars_sharpness} (\textit{right}) shows a broad region of stable performance, with $\gamma_e{=}6$ and $\gamma_\ell{=}1$ providing a balanced trade-off between avoiding speculation and allowing realistic delays, demonstrating the robustness of both StreamReady’s readiness behavior and the ARS metric.

\noindent \textbf{Computation Cost and Inference Latency.}
Figure~\ref{fig:compute} compares scalability across models as video length increases. Baseline Qwen-2-VL suffers from rapid latency growth and out-of-memory failures due to full-attention accumulation. Retrieval-based (ReKV) and activation-based (StreamBridge) methods improve efficiency but still incur rising overhead: ReKV from expanding KV caches and StreamBridge from repeated costly token-compression steps and heavy activation. In contrast, StreamReady maintains stable latency and memory by using a fixed-size compact memory of centroids and prototypes, keeping retrieval cost minimum. Its readiness mechanism, implemented with a single \texttt{<RDY>} token and lightweight MLP head, adds no extra inference overhead, enabling smooth scalability for long-horizon, readiness-aware streaming.

\vspace{-8pt}
\section{Conclusion}
\label{sec:conclusion}
\vspace{-4pt}
We present readiness-aware streaming video understanding, a formulation that evaluates not only \emph{what} a model answers but also \emph{when}, relative to visual evidence. 
To capture this behavior, we propose the Answer Readiness Score (ARS), a timing-aware metric with asymmetric early and late penalties. Building on this formulation, our proposed framework \textbf{StreamReady} integrates long-horizon temporal reasoning with a lightweight readiness mechanism to decide when sufficient evidence has appeared. For evaluation, we introduce \textbf{ProReady-QA}, a long streaming benchmark with proactive multi-turn questions and annotated answer evidence windows. Together, these contributions establish readiness-aware streaming as a step toward models that answer both accurately and on time.

\section{Acknowledgement}
This research is based upon work primarily supported by the National Science Foundation under Grant No. $2331319$. Any opinions, findings, and conclusions or recommendations contained herein are those of the authors and should not be interpreted as necessarily representing the views, either expressed or implied, of the National Science Foundation.

{
    \small
    \bibliographystyle{ieeenat_fullname}
    \bibliography{main}

\begin{thebibliography}{64}
\providecommand{\natexlab}[1]{#1}
\providecommand{\url}[1]{\texttt{#1}}
\expandafter\ifx\csname urlstyle\endcsname\relax
  \providecommand{\doi}[1]{doi: #1}\else
  \providecommand{\doi}{doi: \begingroup \urlstyle{rm}\Url}\fi

\bibitem[Abdin et~al.(2024)Abdin, Jacobs, Awan, Aneja, Awadallah, Awadalla, Bach, Bahree, Bakhtiari, Behl, et~al.]{abdin2024phi}
Marah Abdin, Sam~Ade Jacobs, Ammar~Ahmad Awan, Jyoti Aneja, Ahmed Awadallah, Hany Awadalla, Nguyen Bach, Amit Bahree, Arash Bakhtiari, Harkirat Behl, et~al.
\newblock Phi-3 technical report: A highly capable language model locally on your phone.
\newblock \emph{arXiv preprint arXiv:2404.14219}, 2024.

\bibitem[Alayrac et~al.(2022)Alayrac, Donahue, Luc, Miech, Barr, Hasson, Lenc, Mensch, Millican, Reynolds, et~al.]{alayrac2022flamingo}
Jean-Baptiste Alayrac, Jeff Donahue, Pauline Luc, Antoine Miech, Iain Barr, Yana Hasson, Karel Lenc, Arthur Mensch, Katherine Millican, Malcolm Reynolds, et~al.
\newblock Flamingo: A visual language model for few-shot learning.
\newblock In \emph{Advances in Neural Information Processing Systems}, pages 23716--23736, 2022.

\bibitem[Azad and Rawat(2025)]{azad2025disenq}
Shehreen Azad and Yogesh~Singh Rawat.
\newblock Disenq: Disentangling q-former for activity-biometrics.
\newblock In \emph{Proceedings of the IEEE/CVF International Conference on Computer Vision}, pages 13502--13512, 2025.

\bibitem[Azad et~al.(2025)Azad, Vineet, and Rawat]{azad2025hierarq}
Shehreen Azad, Vibhav Vineet, and Yogesh~Singh Rawat.
\newblock Hierarq: Task-aware hierarchical q-former for enhanced video understanding.
\newblock In \emph{CVPR}, pages 8545--8556, 2025.

\bibitem[Balazevic et~al.(2024)Balazevic, Shi, Papalampidi, Chaabouni, Koppula, and H{\'e}naff]{balazevic2024memory}
Ivana Balazevic, Yuge Shi, Pinelopi Papalampidi, Rahma Chaabouni, Skanda Koppula, and Olivier~J H{\'e}naff.
\newblock Memory consolidation enables long-context video understanding.
\newblock In \emph{Forty-first International Conference on Machine Learning}, 2024.

\bibitem[Chen et~al.(2023)Chen, Zheng, Wang, Xu, Huang, Pan, Wang, Wang, Qiao, Lu, et~al.]{chen2023videollm}
Guo Chen, Yin-Dong Zheng, Jiahao Wang, Jilan Xu, Yifei Huang, Junting Pan, Yi Wang, Yali Wang, Yu Qiao, Tong Lu, et~al.
\newblock Videollm: Modeling video sequence with large language models.
\newblock \emph{arXiv preprint arXiv:2305.13292}, 2023.

\bibitem[Chen et~al.(2024{\natexlab{a}})Chen, Lv, Wu, Lin, Song, Gao, Liu, Gao, Mao, and Shou]{chen2024videollmonline}
Joya Chen, Zhaoyang Lv, Shiwei Wu, Kevin~Qinghong Lin, Chenan Song, Difei Gao, Jia-Wei Liu, Ziteng Gao, Dongxing Mao, and Mike~Zheng Shou.
\newblock Videollm-online: Online video large language model for streaming video.
\newblock In \emph{CVPR}, pages 18407--18418, 2024{\natexlab{a}}.

\bibitem[Chen et~al.(2024{\natexlab{b}})Chen, Wang, Tian, Ye, Gao, Cui, Tong, Hu, Luo, Ma, et~al.]{chen2024internvl2}
Zhe Chen, Weiyun Wang, Hao Tian, Shenglong Ye, Zhangwei Gao, Erfei Cui, Wenwen Tong, Kongzhi Hu, Jiapeng Luo, Zheng Ma, et~al.
\newblock How far are we to gpt-4v? closing the gap to commercial multimodal models with open-source suites.
\newblock \emph{Science China Information Sciences}, 67\penalty0 (12):\penalty0 220101, 2024{\natexlab{b}}.

\bibitem[Cheng et~al.(2024)Cheng, Li, Liu, Guo, Jiang, Liu, Chen, and Zhao]{cheng2024enhancing}
Dingxin Cheng, Mingda Li, Jingyu Liu, Yongxin Guo, Bin Jiang, Qingbin Liu, Xi Chen, and Bo Zhao.
\newblock Enhancing long video understanding via hierarchical event-based memory.
\newblock \emph{arXiv preprint arXiv:2409.06299}, 2024.

\bibitem[Di et~al.(2025)Di, Yu, Zhang, Li, Zhong, Cheng, Li, He, Shu, and Jiang]{di2025rekv}
Shangzhe Di, Zhelun Yu, Guanghao Zhang, Haoyuan Li, Tao Zhong, Hao Cheng, Bolin Li, Wanggui He, Fangxun Shu, and Hao Jiang.
\newblock Streaming video question-answering with in-context video kv-cache retrieval.
\newblock In \emph{ICLR}, 2025.

\bibitem[Fu et~al.(2025{\natexlab{a}})Fu, Dai, Luo, Li, Ren, Zhang, Wang, Zhou, Shen, Zhang, et~al.]{fu2024videomme}
Chaoyou Fu, Yuhan Dai, Yongdong Luo, Lei Li, Shuhuai Ren, Renrui Zhang, Zihan Wang, Chenyu Zhou, Yunhang Shen, Mengdan Zhang, et~al.
\newblock Video-mme: The first-ever comprehensive evaluation benchmark of multi-modal llms in video analysis.
\newblock In \emph{CVPR}, 2025{\natexlab{a}}.

\bibitem[Fu et~al.(2025{\natexlab{b}})Fu, Lin, Wang, Zhang, Shen, Liu, Cao, Long, Gao, Li, et~al.]{fu2025vita}
Chaoyou Fu, Haojia Lin, Xiong Wang, Yi-Fan Zhang, Yunhang Shen, Xiaoyu Liu, Haoyu Cao, Zuwei Long, Heting Gao, Ke Li, et~al.
\newblock Vita-1.5: Towards gpt-4o level real-time vision and speech interaction.
\newblock \emph{arXiv preprint arXiv:2501.01957}, 2025{\natexlab{b}}.

\bibitem[Fu et~al.(2025{\natexlab{c}})Fu, Yang, Li, Peng, Lin, Wei, Hu, Xie, and Zheng]{fu2025vispeak}
Shenghao Fu, Qize Yang, Yuan-Ming Li, Yi-Xing Peng, Kun-Yu Lin, Xihan Wei, Jian-Fang Hu, Xiaohua Xie, and Wei-Shi Zheng.
\newblock Vispeak: Visual instruction feedback in streaming videos.
\newblock In \emph{ICCV}, 2025{\natexlab{c}}.

\bibitem[Grauman et~al.(2022)Grauman, Westbury, Byrne, Chavis, Furnari, Girdhar, Hamburger, Jiang, Liu, Liu, et~al.]{grauman2022ego4d}
Kristen Grauman, Andrew Westbury, Eugene Byrne, Zachary Chavis, Antonino Furnari, Rohit Girdhar, Jackson Hamburger, Hao Jiang, Miao Liu, Xingyu Liu, et~al.
\newblock Ego4d: Around the world in 3,000 hours of egocentric video.
\newblock In \emph{CVPR}, pages 18995--19012, 2022.

\bibitem[He et~al.(2024)He, Li, Jang, Jia, Cao, Shah, Shrivastava, and Lim]{he2024ma}
Bo He, Hengduo Li, Young~Kyun Jang, Menglin Jia, Xuefei Cao, Ashish Shah, Abhinav Shrivastava, and Ser-Nam Lim.
\newblock Ma-lmm: Memory-augmented large multimodal model for long-term video understanding.
\newblock In \emph{CVPR}, pages 13504--13514, 2024.

\bibitem[Huang et~al.(2024)Huang, Liao, Radhakrishnan, Yin, Molchanov, Yu, and Kautz]{huang2024lita}
De-An Huang, Shijia Liao, Subhashree Radhakrishnan, Hongxu Yin, Pavlo Molchanov, Zhiding Yu, and Jan Kautz.
\newblock Lita: Language instructed temporal-localization assistant.
\newblock In \emph{ECCV}, pages 202--218. Springer, 2024.

\bibitem[Huang et~al.(2020)Huang, Xiong, Rao, Wang, and Lin]{huang2020movienet}
Qingqiu Huang, Yu Xiong, Anyi Rao, Jiaze Wang, and Dahua Lin.
\newblock Movienet: A holistic dataset for movie understanding.
\newblock In \emph{ECCV}, pages 709--727. Springer, 2020.

\bibitem[Huang et~al.(2025)Huang, Li, Li, Wang, Zeng, Liang, Wu, Chen, Li, and Wang]{huang2024videochatonline}
Zhenpeng Huang, Xinhao Li, Jiaqi Li, Jing Wang, Xiangyu Zeng, Cheng Liang, Tao Wu, Xi Chen, Liang Li, and Limin Wang.
\newblock Online video understanding: Ovbench and videochat-online.
\newblock In \emph{CVPR}, pages 3328--3338, 2025.

\bibitem[Kim et~al.(2025)Kim, Shim, Choi, and Chang]{kim2025infinipot}
Minsoo Kim, Kyuhong Shim, Jungwook Choi, and Simyung Chang.
\newblock Infinipot-v: Memory-constrained kv cache compression for streaming video understanding.
\newblock In \emph{NeurIPS}, 2025.

\bibitem[Kugo et~al.(2025)Kugo, Li, Li, Gupta, Khatua, Jain, Patel, Kyuragi, Ishii, Tanabiki, et~al.]{kugo2025videomultiagents}
Noriyuki Kugo, Xiang Li, Zixin Li, Ashish Gupta, Arpandeep Khatua, Nidhish Jain, Chaitanya Patel, Yuta Kyuragi, Yasunori Ishii, Masamoto Tanabiki, et~al.
\newblock Videomultiagents: A multi-agent framework for video question answering.
\newblock \emph{arXiv preprint arXiv:2504.20091}, 2025.

\bibitem[Li et~al.(2024{\natexlab{a}})Li, Zhang, Guo, Zhang, Li, Zhang, Zhang, Li, Liu, and Li]{li2024llava}
Bo Li, Yuanhan Zhang, Dong Guo, Renrui Zhang, Feng Li, Hao Zhang, Kaichen Zhang, Yanwei Li, Ziwei Liu, and Chunyuan Li.
\newblock Llava-onevision: Easy visual task transfer.
\newblock \emph{arXiv preprint arXiv:2408.03326}, 2024{\natexlab{a}}.

\bibitem[Li et~al.(2024{\natexlab{b}})Li, Zhang, Guo, Zhang, Li, Zhang, Zhang, Zhang, Li, Liu, et~al.]{li2024llavaov}
Bo Li, Yuanhan Zhang, Dong Guo, Renrui Zhang, Feng Li, Hao Zhang, Kaichen Zhang, Peiyuan Zhang, Yanwei Li, Ziwei Liu, et~al.
\newblock Llava-onevision: Easy visual task transfer.
\newblock \emph{arXiv preprint arXiv:2408.03326}, 2024{\natexlab{b}}.

\bibitem[Li et~al.(2024{\natexlab{c}})Li, Wang, He, Li, Wang, Liu, Wang, Xu, Chen, Luo, et~al.]{li2024mvbench}
Kunchang Li, Yali Wang, Yinan He, Yizhuo Li, Yi Wang, Yi Liu, Zun Wang, Jilan Xu, Guo Chen, Ping Luo, et~al.
\newblock Mvbench: A comprehensive multi-modal video understanding benchmark.
\newblock In \emph{CVPR}, pages 22195--22206, 2024{\natexlab{c}}.

\bibitem[Li et~al.(2025{\natexlab{a}})Li, Li, Wang, He, Wang, Wang, and Qiao]{li2025videomamba}
Kunchang Li, Xinhao Li, Yi Wang, Yinan He, Yali Wang, Limin Wang, and Yu Qiao.
\newblock Videomamba: State space model for efficient video understanding.
\newblock In \emph{European Conference on Computer Vision}, pages 237--255. Springer, 2025{\natexlab{a}}.

\bibitem[Li et~al.(2024{\natexlab{d}})Li, Wang, Yu, Zeng, Zhu, Huang, Gao, Li, He, Wang, et~al.]{li2024videochatflash}
Xinhao Li, Yi Wang, Jiashuo Yu, Xiangyu Zeng, Yuhan Zhu, Haian Huang, Jianfei Gao, Kunchang Li, Yinan He, Chenting Wang, et~al.
\newblock Videochat-flash: Hierarchical compression for long-context video modeling.
\newblock \emph{arXiv preprint arXiv:2501.00574}, 2024{\natexlab{d}}.

\bibitem[Li et~al.(2024{\natexlab{e}})Li, Wang, and Jia]{li2024llamavid}
Yanwei Li, Chengyao Wang, and Jiaya Jia.
\newblock Llama-vid: An image is worth 2 tokens in large language models.
\newblock In \emph{ECCV}, pages 323--340. Springer, 2024{\natexlab{e}}.

\bibitem[Li et~al.(2025{\natexlab{b}})Li, Niu, Miao, Ge, Zhou, He, Dong, Duan, Ding, Qian, et~al.]{li2025ovobench}
Yifei Li, Junbo Niu, Ziyang Miao, Chunjiang Ge, Yuanhang Zhou, Qihao He, Xiaoyi Dong, Haodong Duan, Shuangrui Ding, Rui Qian, et~al.
\newblock Ovo-bench: How far is your video-llms from real-world online video understanding?
\newblock \emph{arXiv preprint arXiv:2501.05510}, 2025{\natexlab{b}}.

\bibitem[Lin et~al.(2024)Lin, Fang, Chen, Wan, Luo, Li, Liu, and Sun]{lin2024streamingbench}
Junming Lin, Zheng Fang, Chi Chen, Zihao Wan, Fuwen Luo, Peng Li, Yang Liu, and Maosong Sun.
\newblock Streamingbench: Assessing the gap for mllms to achieve streaming video understanding.
\newblock \emph{arXiv preprint arXiv:2411.03628}, 2024.

\bibitem[Maaz et~al.(2024)Maaz, Rasheed, Khan, and Khan]{maaz-etal-2024-video}
Muhammad Maaz, Hanoona Rasheed, Salman Khan, and Fahad Khan.
\newblock Video-{C}hat{GPT}: Towards detailed video understanding via large vision and language models.
\newblock In \emph{Proceedings of the 62nd Annual Meeting of the Association for Computational Linguistics}, 2024.

\bibitem[Mangalam et~al.(2023)Mangalam, Akshulakov, and Malik]{mangalam2023egoschema}
Karttikeya Mangalam, Raiymbek Akshulakov, and Jitendra Malik.
\newblock Egoschema: A diagnostic benchmark for very long-form video language understanding.
\newblock \emph{Advances in Neural Information Processing Systems}, 36:\penalty0 46212--46244, 2023.

\bibitem[Montes and Lozano(2025)]{montes2025viqagent}
Tony Montes and Fernando Lozano.
\newblock Viqagent: Zero-shot video question answering via agent with open-vocabulary grounding validation.
\newblock \emph{arXiv preprint arXiv:2505.15928}, 2025.

\bibitem[Qian et~al.(2024)Qian, Dong, Zhang, Zang, Ding, Lin, and Wang]{qian2024streaming}
Rui Qian, Xiaoyi Dong, Pan Zhang, Yuhang Zang, Shuangrui Ding, Dahua Lin, and Jiaqi Wang.
\newblock Streaming long video understanding with large language models.
\newblock \emph{Advances in Neural Information Processing Systems}, 37:\penalty0 119336--119360, 2024.

\bibitem[Qian et~al.(2025)Qian, Ding, Dong, Zhang, Zang, Cao, Lin, and Wang]{qian2025dispider}
Rui Qian, Shuangrui Ding, Xiaoyi Dong, Pan Zhang, Yuhang Zang, Yuhang Cao, Dahua Lin, and Jiaqi Wang.
\newblock Dispider: Enabling video llms with active real-time interaction via disentangled perception, decision, and reaction.
\newblock In \emph{CVPR}, 2025.

\bibitem[Ren et~al.(2024)Ren, Yao, Li, Sun, and Hou]{ren2024timechat}
Shuhuai Ren, Linli Yao, Shicheng Li, Xu Sun, and Lu Hou.
\newblock Timechat: A time-sensitive multimodal large language model for long video understanding.
\newblock In \emph{Proceedings of the IEEE/CVF Conference on Computer Vision and Pattern Recognition}, pages 14313--14323, 2024.

\bibitem[Schiappa et~al.(2024{\natexlab{a}})Schiappa, Abdullah, Azad, Claypoole, Cogswell, Divakaran, and Rawat]{schiappa2024probing}
Madeline Schiappa, Raiyaan Abdullah, Shehreen Azad, Jared Claypoole, Michael Cogswell, Ajay Divakaran, and Yogesh Rawat.
\newblock Probing conceptual understanding of large visual-language models.
\newblock In \emph{Proceedings of the IEEE/CVF Conference on Computer Vision and Pattern Recognition}, pages 1797--1807, 2024{\natexlab{a}}.

\bibitem[Schiappa et~al.(2024{\natexlab{b}})Schiappa, Azad, Vs, Ge, Miksik, Rawat, and Vineet]{schiappa2024robustness}
Madeline~Chantry Schiappa, Shehreen Azad, Sachidanand Vs, Yunhao Ge, Ondrej Miksik, Yogesh~S Rawat, and Vibhav Vineet.
\newblock Robustness analysis on foundational segmentation models.
\newblock In \emph{Proceedings of the IEEE/CVF Conference on Computer Vision and Pattern Recognition}, pages 1786--1796, 2024{\natexlab{b}}.

\bibitem[Shen et~al.(2024)Shen, Xiong, Zhao, Wu, Chen, Zhu, Liu, Xiao, Varadarajan, Bordes, et~al.]{shen2024longvu}
Xiaoqian Shen, Yunyang Xiong, Changsheng Zhao, Lemeng Wu, Jun Chen, Chenchen Zhu, Zechun Liu, Fanyi Xiao, Balakrishnan Varadarajan, Florian Bordes, et~al.
\newblock Longvu: Spatiotemporal adaptive compression for long video-language understanding.
\newblock \emph{arXiv preprint arXiv:2410.17434}, 2024.

\bibitem[Song et~al.(2024{\natexlab{a}})Song, Chai, Wang, Zhang, Zhou, Wu, Chi, Guo, Ye, Zhang, et~al.]{song2024moviechat}
Enxin Song, Wenhao Chai, Guanhong Wang, Yucheng Zhang, Haoyang Zhou, Feiyang Wu, Haozhe Chi, Xun Guo, Tian Ye, Yanting Zhang, et~al.
\newblock Moviechat: From dense token to sparse memory for long video understanding.
\newblock In \emph{Proceedings of the IEEE/CVF Conference on Computer Vision and Pattern Recognition}, pages 18221--18232, 2024{\natexlab{a}}.

\bibitem[Song et~al.(2024{\natexlab{b}})Song, Chai, Ye, Hwang, Li, and Wang]{song2024moviechat+}
Enxin Song, Wenhao Chai, Tian Ye, Jenq-Neng Hwang, Xi Li, and Gaoang Wang.
\newblock Moviechat+: Question-aware sparse memory for long video question answering.
\newblock \emph{arXiv preprint arXiv:2404.17176}, 2024{\natexlab{b}}.

\bibitem[Wang et~al.(2025{\natexlab{a}})Wang, Feng, Lai, Xu, Li, Ge, Dehghan, Cao, and Huang]{wang2025streambridge}
Haibo Wang, Bo Feng, Zhengfeng Lai, Mingze Xu, Shiyu Li, Weifeng Ge, Afshin Dehghan, Meng Cao, and Ping Huang.
\newblock Streambridge: Turning your offline video large language model into a proactive streaming assistant.
\newblock In \emph{NeurIPS}, 2025{\natexlab{a}}.

\bibitem[Wang et~al.(2024{\natexlab{a}})Wang, Bai, Tan, Wang, Fan, Bai, Chen, Liu, Wang, Ge, et~al.]{qwen2vl}
Peng Wang, Shuai Bai, Sinan Tan, Shijie Wang, Zhihao Fan, Jinze Bai, Keqin Chen, Xuejing Liu, Jialin Wang, Wenbin Ge, et~al.
\newblock Qwen2-vl: Enhancing vision-language model's perception of the world at any resolution.
\newblock \emph{arXiv preprint arXiv:2409.12191}, 2024{\natexlab{a}}.

\bibitem[Wang et~al.(2024{\natexlab{b}})Wang, Zhang, Zohar, and Yeung-Levy]{wang2024videoagent}
Xiaohan Wang, Yuhui Zhang, Orr Zohar, and Serena Yeung-Levy.
\newblock Videoagent: Long-form video understanding with large language model as agent.
\newblock In \emph{European Conference on Computer Vision}, pages 58--76. Springer, 2024{\natexlab{b}}.

\bibitem[Wang et~al.(2024{\natexlab{c}})Wang, Meng, Wang, Liang, Wei, Zhang, and Zhao]{wang2024videollm}
Yueqian Wang, Xiaojun Meng, Yuxuan Wang, Jianxin Liang, Jiansheng Wei, Huishuai Zhang, and Dongyan Zhao.
\newblock Videollm knows when to speak: Enhancing time-sensitive video comprehension with video-text duet interaction format.
\newblock \emph{arXiv preprint arXiv:2411.17991}, 2024{\natexlab{c}}.

\bibitem[Wang et~al.(2025{\natexlab{b}})Wang, Meng, Wang, Zhang, and Zhao]{wang2025proactivevideoqacomprehensivebenchmarkevaluating}
Yueqian Wang, Xiaojun Meng, Yifan Wang, Huishuai Zhang, and Dongyan Zhao.
\newblock Proactivevideoqa: A comprehensive benchmark evaluating proactive interactions in video large language models, 2025{\natexlab{b}}.

\bibitem[Wang et~al.(2025{\natexlab{c}})Wang, Wang, Chen, Wu, Zhao, and Zheng]{wang2025omnimmi}
Yuxuan Wang, Yueqian Wang, Bo Chen, Tong Wu, Dongyan Zhao, and Zilong Zheng.
\newblock Omnimmi: A comprehensive multi-modal interaction benchmark in streaming video contexts.
\newblock In \emph{CVPR}, pages 18925--18935, 2025{\natexlab{c}}.

\bibitem[Wang et~al.(2025{\natexlab{d}})Wang, Xie, Liu, and Zheng]{wang2024videollamb}
Yuxuan Wang, Cihang Xie, Yang Liu, and Zilong Zheng.
\newblock Videollamb: Long-context video understanding with recurrent memory bridges.
\newblock In \emph{ICCV}, 2025{\natexlab{d}}.

\bibitem[Wang et~al.(2025{\natexlab{e}})Wang, Yu, Stengel-Eskin, Yoon, Cheng, Bertasius, and Bansal]{wang2025videotree}
Ziyang Wang, Shoubin Yu, Elias Stengel-Eskin, Jaehong Yoon, Feng Cheng, Gedas Bertasius, and Mohit Bansal.
\newblock Videotree: Adaptive tree-based video representation for llm reasoning on long videos.
\newblock In \emph{Proceedings of the Computer Vision and Pattern Recognition Conference}, pages 3272--3283, 2025{\natexlab{e}}.

\bibitem[Xiong et~al.(2025)Xiong, Yang, Yu, Zhuge, Zhang, Zhu, and Lu]{xiong2025streamchat}
Haomiao Xiong, Zongxin Yang, Jiazuo Yu, Yunzhi Zhuge, Lu Zhang, Jiawen Zhu, and Huchuan Lu.
\newblock Streaming video understanding and multi-round interaction with memory-enhanced knowledge.
\newblock \emph{arXiv preprint arXiv:2501.13468}, 2025.

\bibitem[Xu et~al.(2017)Xu, Zhao, Xiao, Wu, Zhang, He, and Zhuang]{xu2017video}
Dejing Xu, Zhou Zhao, Jun Xiao, Fei Wu, Hanwang Zhang, Xiangnan He, and Yueting Zhuang.
\newblock Video question answering via gradually refined attention over appearance and motion.
\newblock In \emph{Proceedings of the 25th ACM international conference on Multimedia}, pages 1645--1653, 2017.

\bibitem[Yang et~al.(2025{\natexlab{a}})Yang, Tang, Zhao, An, Hu, Li, Zhuang, Wang, Lu, Zhang, et~al.]{yang2025streamagent}
Haolin Yang, Feilong Tang, Linxiao Zhao, Xiang An, Ming Hu, Huifa Li, Xinlin Zhuang, Boqian Wang, Yifan Lu, Xiaofeng Zhang, et~al.
\newblock Streamagent: Towards anticipatory agents for streaming video understanding.
\newblock \emph{arXiv preprint arXiv:2508.01875}, 2025{\natexlab{a}}.

\bibitem[Yang et~al.(2025{\natexlab{b}})Yang, Zhao, Shukla, Singh, Mishra, Zhang, and Ren]{yang2025streammem}
Yanlai Yang, Zhuokai Zhao, Satya~Narayan Shukla, Aashu Singh, Shlok~Kumar Mishra, Lizhu Zhang, and Mengye Ren.
\newblock Streammem: Query-agnostic kv cache memory for streaming video understanding.
\newblock \emph{arXiv preprint arXiv:2508.15717}, 2025{\natexlab{b}}.

\bibitem[Yao et~al.(2025)Yao, Li, Wei, Li, Ren, Liu, Ouyang, Wang, Li, Li, et~al.]{yao2025timechat}
Linli Yao, Yicheng Li, Yuancheng Wei, Lei Li, Shuhuai Ren, Yuanxin Liu, Kun Ouyang, Lean Wang, Shicheng Li, Sida Li, et~al.
\newblock Timechat-online: 80\% visual tokens are naturally redundant in streaming videos.
\newblock \emph{arXiv preprint arXiv:2504.17343}, 2025.

\bibitem[Yao et~al.(2024)Yao, Yu, Zhang, Wang, Cui, Zhu, Cai, Li, Zhao, He, et~al.]{yao2024minicpmv}
Yuan Yao, Tianyu Yu, Ao Zhang, Chongyi Wang, Junbo Cui, Hongji Zhu, Tianchi Cai, Haoyu Li, Weilin Zhao, Zhihui He, et~al.
\newblock Minicpm-v: A gpt-4v level mllm on your phone.
\newblock \emph{arXiv preprint arXiv:2408.01800}, 2024.

\bibitem[Yu et~al.(2019)Yu, Xu, Yu, Yu, Zhao, Zhuang, and Tao]{yu2019activitynet}
Zhou Yu, Dejing Xu, Jun Yu, Ting Yu, Zhou Zhao, Yueting Zhuang, and Dacheng Tao.
\newblock Activitynet-qa: A dataset for understanding complex web videos via question answering.
\newblock In \emph{Proceedings of the AAAI Conference on Artificial Intelligence}, pages 9127--9134, 2019.

\bibitem[Zeng et~al.(2025)Zeng, Qiu, Zhang, Li, Wang, Li, Yan, Tian, Tian, Zhao, Wang, and Wang]{zeng2025streamforest}
Xiangyu Zeng, Kefan Qiu, Qingyu Zhang, Xinhao Li, Jing Wang, Jiaxin Li, Ziang Yan, Kun Tian, Meng Tian, Xinhai Zhao, Yi Wang, and Limin Wang.
\newblock Streamforest: Efficient online video understanding with persistent event memory, 2025.

\bibitem[Zhang et~al.(2025{\natexlab{a}})Zhang, Hannan, Kleiner, Aydemir, Xie, Lan, Seidl, Tresp, and Gu]{zhang2025avila}
Gengyuan Zhang, Tanveer Hannan, Hermine Kleiner, Beste Aydemir, Xinyu Xie, Jian Lan, Thomas Seidl, Volker Tresp, and Jindong Gu.
\newblock Avila: Asynchronous vision-language agent for streaming multimodal data interaction.
\newblock \emph{arXiv preprint arXiv:2506.18472}, 2025{\natexlab{a}}.

\bibitem[Zhang et~al.(2024{\natexlab{a}})Zhang, Wang, Tang, Liu, Feng, Dai, and Jin]{zhang2024flash}
Haoji Zhang, Yiqin Wang, Yansong Tang, Yong Liu, Jiashi Feng, Jifeng Dai, and Xiaojie Jin.
\newblock Flash-vstream: Memory-based real-time understanding for long video streams.
\newblock \emph{arXiv preprint arXiv:2406.08085}, 2024{\natexlab{a}}.

\bibitem[Zhang et~al.(2025{\natexlab{b}})Zhang, Wang, Tang, Liu, Feng, Dai, and Jin]{zhang2024flashvstream}
Haoji Zhang, Yiqin Wang, Yansong Tang, Yong Liu, Jiashi Feng, Jifeng Dai, and Xiaojie Jin.
\newblock Flash-vstream: Memory-based real-time understanding for long video streams.
\newblock In \emph{ICCV}, 2025{\natexlab{b}}.

\bibitem[Zhang et~al.(2024{\natexlab{b}})Zhang, Zhang, Li, Zeng, Yang, Zhang, Wang, Tan, Li, and Liu]{zhang2024longva}
Peiyuan Zhang, Kaichen Zhang, Bo Li, Guangtao Zeng, Jingkang Yang, Yuanhan Zhang, Ziyue Wang, Haoran Tan, Chunyuan Li, and Ziwei Liu.
\newblock Long context transfer from language to vision.
\newblock \emph{arXiv preprint arXiv:2406.16852}, 2024{\natexlab{b}}.

\bibitem[Zhang et~al.(2024{\natexlab{c}})Zhang, Li, Liu, Lee, Gui, Fu, Feng, Liu, and Li]{zhang2024llavanextvideo}
Yuanhan Zhang, Bo Li, haotian Liu, Yong~jae Lee, Liangke Gui, Di Fu, Jiashi Feng, Ziwei Liu, and Chunyuan Li.
\newblock Llava-next: A strong zero-shot video understanding model, 2024{\natexlab{c}}.

\bibitem[Zhang et~al.(2024{\natexlab{d}})Zhang, Wu, Li, Li, Ma, Liu, and Li]{zhang2024llavavideo}
Yuanhan Zhang, Jinming Wu, Wei Li, Bo Li, Zejun Ma, Ziwei Liu, and Chunyuan Li.
\newblock Video instruction tuning with synthetic data.
\newblock \emph{arXiv preprint arXiv:2410.02713}, 2024{\natexlab{d}}.

\bibitem[Zhi et~al.(2025)Zhi, Wu, Li, Li, Shao, Zhou, et~al.]{zhi2025videoagent2}
Zhuo Zhi, Qiangqiang Wu, Wenbo Li, Yinchuan Li, Kun Shao, Kaiwen Zhou, et~al.
\newblock Videoagent2: Enhancing the llm-based agent system for long-form video understanding by uncertainty-aware cot.
\newblock \emph{arXiv preprint arXiv:2504.04471}, 2025.

\bibitem[Zhou et~al.(2024)Zhou, Shu, Zhao, Wu, Xiao, Yang, Xiong, Zhang, Huang, and Liu]{zhou2024mlvu}
Junjie Zhou, Yan Shu, Bo Zhao, Boya Wu, Shitao Xiao, Xi Yang, Yongping Xiong, Bo Zhang, Tiejun Huang, and Zheng Liu.
\newblock Mlvu: A comprehensive benchmark for multi-task long video understanding.
\newblock \emph{arXiv preprint arXiv:2406.04264}, 2024.

\bibitem[Zou et~al.(2024)Zou, Luo, Xie, Lv, Wang, Chen, Wang, Zhang, Zhang, et~al.]{zou2024seconds}
Heqing Zou, Tianze Luo, Guiyang Xie, Fengmao Lv, Guangcong Wang, Junyang Chen, Zhuochen Wang, Hansheng Zhang, Huaijian Zhang, et~al.
\newblock From seconds to hours: Reviewing multimodal large language models on comprehensive long video understanding.
\newblock \emph{arXiv preprint arXiv:2409.18938}, 2024.

\end{thebibliography}
}

\clearpage
\setcounter{page}{1}
\renewcommand{\thesection}{\Alph{section}}
\setcounter{section}{0}

\maketitlesupplementary
 
In this supplementary material, we provide additional ablation studies in \S \ref{sec:aba_sup}, followed by additional quantitative and qualitative analysis in \S \ref{sec:res_qual}. Next, we provide the details of ProReady-QA generation pipeline in \S \ref{sec:generation_supp}, followed by additional implementation details in \S \ref{sec:imp_sup}. Finally, we outline some directions for future research in \S \ref{sec:future}.

\section{Additional Ablation}
\label{sec:aba_sup}
We perform additional ablation studies on 3 tasks of ProReady-QA (REC, GSD, CTD), 2 tasks of StreamingBench (real-time perception and contextual reasoning), and VideoMME (long) to show the generalization of our framework on both proactive, and non-proactive streaming, and offline long-video benchmarks.

\noindent\textbf{Contribution of Each Memory Bank.}
Table \ref{tab:abla_sup} (\textit{top}) shows that short-term memory alone is insufficient for long streaming videos, since it captures only local context and quickly loses the long-range cues needed for temporal aggregation and causal reasoning. Adding long-term memory, thus, brings substantial gains across all datasets by capturing broader temporal context. The largest improvement comes from contextual memory, which allows the model to reuse prior QA information, maintain cross-turn consistency, and exploit long-range dependencies essential for multi-turn dialogue. This is especially evident on StreamingBench’s contextual reasoning task and in the ARS gains on ProReady-QA, where stronger evidence modeling improves both accuracy and timing. While, VideoMME-Long benefits from long-term memory, contextual memory/reasoning is irrelevant for this benchmark due to its single-question format. Overall, combining hierarchical visual memory tree with lightweight contextual memory is key for StreamReady's reliable multi-turn reasoning in long-stream settings.

\noindent\textbf{Design Choice of Visual Memory Tree.}
Existing long-term memory strategies typically rely on either similarity-based grouping \cite{azad2025hierarq, song2024moviechat} or caption-based summarization \cite{zeng2025streamforest, xiong2025streamchat} to control memory growth and preserve information. Table \ref{tab:abla_sup} (\textit{middle}) compares these approaches in both flat and hierarchical settings. Flat variants of both methods perform poorly, since collapsing all information into a single pool erases fine-grained cues needed for detailed evidence retrieval in long videos. Adding hierarchy alleviates this by organizing memory into increasingly abstract levels, reducing tokens while retaining the subtle patterns essential for long-form reasoning. Within this structure, similarity-based clustering consistently outperforms caption-based summarization, which is prone to semantic drift and incurs higher compute and latency costs. Our adaptive hierarchical clustering extends similarity-based designs across multiple abstraction levels, preserving temporal structure while keeping memory compact. This yields the strongest accuracy and ARS on ProReady-QA and the best performance on StreamingBench and VideoMME-Long, demonstrating the effectiveness of adaptive coarse-to-fine clustering for scalable long-term memory construction.

\noindent \textbf{Design Choice of Query-Aware Reasoning.}
Table \ref{tab:abla_sup} (\textit{bottom}) analyzes how different forms of query-aware reasoning affect performance.
Without query-aware retrieval, the model cannot focus on the correct memory region or judge evidence sufficiency, leading to low accuracy and ARS. 
Short-term awareness offers modest gains but remains limited because it captures local changes but misses the long-range dependencies needed for long-term comprehension. Adding long-term awareness yields much larger improvements, with centroid-level reasoning consistently outperforming prototype-level reasoning across all datasets. Because centroids preserve finer visual details, it enables more precise evidence selection and better timing; whereas prototype-only reasoning loses cues needed for accurate temporal localization. The coarse-to-fine design of using prototypes for broad context and centroids for detailed refinement achieves the best results, delivering the strongest accuracy and ARS on all ProReady-QA tasks. A similar trend appears in StreamingBench and VideoMME-Long, where coarse-to-fine retrieval supports both global relevance filtering and precise evidence grounding. These results indicate that layered long-term reasoning not only improves overall performance but also enhances answer readiness by enabling more reliable assessment of evidence sufficiency.

\begin{table*}[t!]
    \centering
    \small
    \caption{\textbf{Additional architectural ablation studies} on ProReady-QA, StreamingBench, and VideoMME long.}
    \begin{tabular}{l|cc|cc|cc|cc|c}
    \toprule
         \multirow{3}{*}{\textbf{Method}} & \multicolumn{6}{c|}{\textbf{ProReady-QA}} & \multicolumn{2}{c|}{\textbf{StreamingBench}} &\textbf{Vid-MME}\\
         \cline{2-10}
         &\multicolumn{2}{c|}{\textbf{REC}} & \multicolumn{2}{c|}{\textbf{GSD}} & \multicolumn{2}{c|}{\textbf{CTD}} &  \textbf{Real} & \textbf{Context.} & \textbf{Long}\\
         & Acc. & ARS & Acc. & ARS & Acc. & ARS & Acc. & Acc. & Acc.\\
         \midrule
         \multicolumn{6}{l}{\textcolor{gray}{Contribution of Each Memory Bank}}\\
         Short-Term Memory ($\mathcal{M}_{\text{V1}}$)  &12.7& 0.51 & 24.1 & 0.56 & 31.7 & 0.49 & 41.3 & 12.4 & 36.7\\
         + Long-Term Memory ($\mathcal{M}_{\text{V2}}, \mathcal{M}_{\text{V3}}$) &37.6&0.66 & 55.3 & \textbf{0.68} & 40.7 & 0.58 & 69.4 & 32.1 & \textbf{62.6}\\
         \rowcolor{green!10}
         \textbf{+ Contextual Memory ($\mathbf{\mathcal{M}_{\text{C}}}$) }&\textbf{39.6}&\textbf{0.68}& \textbf{61.2} & \textbf{0.68} & \textbf{43.5} & \textbf{0.59} &\textbf{78.3} & \textbf{48.2} & \textcolor{gray}{N/A} \\
         \midrule
    \multicolumn{6}{l}{\textcolor{gray}{Design choice of Visual Memory Tree}}\\
          Flat w/ similarity \cite{azad2025hierarq}&32.6 & 0.59 &   47.8 & 0.61 & 32.2 & 0.50 & 72.5 & 42.4 & 58.4 \\
          Flat w/ captioner \cite{qian2024streaming}&34.5 & 0.50 & 51.7 & 0.52 & 34.7 & 0.43 & 71.2 & 43.8 & 60.1\\
          Hierarchical  w/ similarity \cite{huang2024videochatonline} & 37.7 & 0.57 & 59.2 & 0.63 & 40.6 & 0.54 & 74.5 & 45.2 & 62.3\\
         Hierarchical w/ captioner \cite{xiong2025streamchat} &35.8 & 0.54 & 60.3 & 0.58 & 39.4 & 0.51 & 72.6 & 44.3 & 62.1  \\
          \rowcolor{green!10}
          \textbf{Hierarchical w/ adaptive clustering}  &\textbf{39.6} &\textbf{0.68}& \textbf{61.2} & \textbf{0.68} & \textbf{43.5} & \textbf{0.59} &\textbf{78.3} & \textbf{48.2} & \textbf{62.6}\\   
          \midrule 
          \multicolumn{6}{l}{\textcolor{gray}{Design Choice of Query-Aware Reasoning}}\\
          No awareness & 29.6 & 0.41 & 44.2 & 0.34 & 22.8 & 0.38 & 64.3 & 37.1 & 53.4\\
          Short-term (ST) aware & 30.1 & 0.41 & 49.6 & 0.37 & 31.8 & 0.39 & 68.2 & 37.5 & 56.2\\
          ST + Centroid-level long-term aware (only $\mathcal{M}_{\text{V2}}$) & 38.1 & 0.67 & {60.7} & 0.63 & 42.7 & 0.57 & 76.8 & 45.1 & 61.9\\ 
          ST + Prototype-level long-term aware (only $\mathcal{M}_{\text{V3}}$) & 32.3 & 0.44 & 52.2 & 0.54 & 39.1 & 0.46 & 64.7 & 41.4 & 59.4\\ 
          \rowcolor{green!10}
          \textbf{ST + Coarse-to-fine long-term aware} & \textbf{39.6} & \textbf{0.68} & \textbf{61.2} & \textbf{0.68} & \textbf{43.5} & \textbf{0.59} &\textbf{78.3} & \textbf{48.2} & \textbf{62.6} \\
          \bottomrule
    \end{tabular}
    \label{tab:abla_sup}
\end{table*}

\begin{figure*}[t!]
    \centering
    \includegraphics[width=\linewidth]{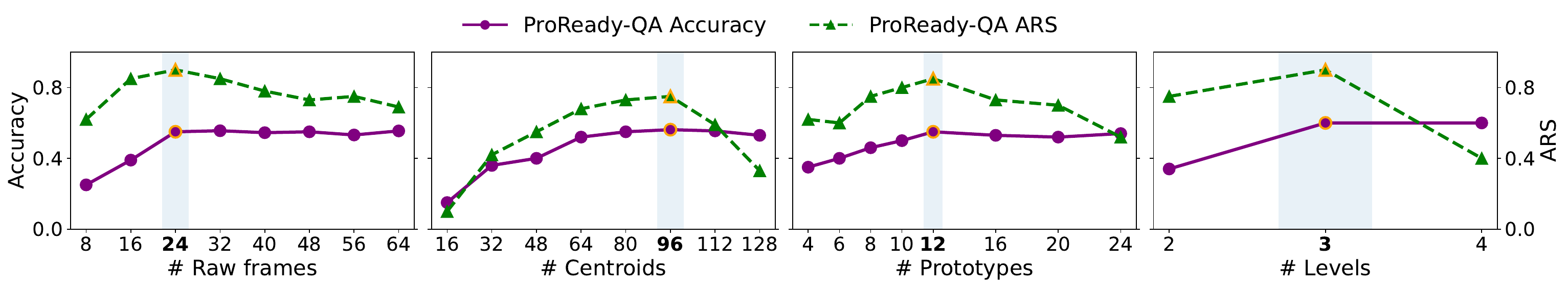}
    \includegraphics[width=\linewidth]{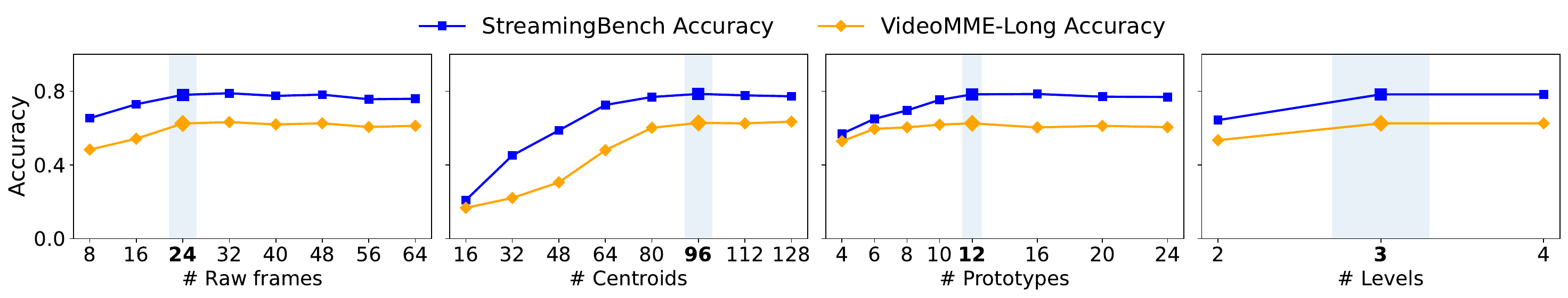}
    \caption{\textbf{Effect of level capacity and depth of visual memory tree on overall performance.} When varying the capacity of a specific level, the capacities of all other levels are held fixed. For the 2-level variant, the centroid and prototype capacities are merged into a single long-term level. For the 4-level variant, the prototype capacity is split across two abstraction layers to create two prototype levels. The selected capacity/depth is highlighted.}
    \label{fig:mem_bank_size_ablation}

\end{figure*}

\section{Additional Analysis and Discussion}
\label{sec:res_qual}

\subsection{Model Behavior Analysis}
\textbf{Effect of Level Capacity and Depth of Visual Memory Tree.} Figure \ref{fig:mem_bank_size_ablation} shows that each level of the visual memory tree has an optimal capacity beyond which accuracy plateaus while ARS declines. For level-1 raw frames, accuracy quickly stabilizes but ARS gradually drops as the FIFO buffer grows, since larger buffers slow retrieval and introduce timing penalties. The effect is sharper for level-2 centroids: moderate cluster counts retain the mid-level temporal detail needed to track subtle event changes across long videos, but larger centroid sets dilute discriminative structure and add latency, causing ARS to drop sharply. This matters across all benchmarks: ProReady-QA requires step-wise, causal, and clue-based evidence; StreamingBench relies on real-time cues and cross-turn context; VideoMME-Long demands precise long-range localization; so losing mid-level detail directly harms reasoning and timing. Level-3 prototypes show a similar pattern: small sets provide useful abstraction, while overly large sets add noise and latency without improving retrieval performance. Varying the depth of the tree further confirms that a 3-level structure balances abstraction and detail best; 2-level variants lose necessary mid-level cues, while 4-level designs over-fragment memory and introduce timing penalties. Overall, a configuration of 24 frames, 96 centroids, and 12 prototypes achieves strong accuracy and stable ARS by encoding long temporal structure efficiently while keeping retrieval fast.

\begin{figure}
    \centering
    \includegraphics[width=.8\linewidth]{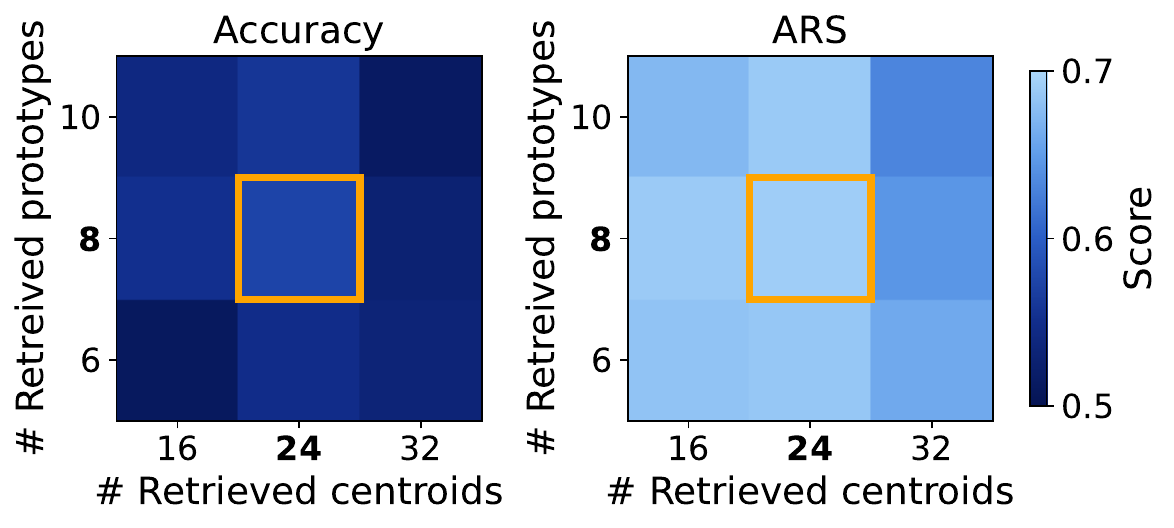}
    \caption{\textbf{Effect of number of query-aware retrieval slots} on ProReady-QA.}
    \label{fig:retrieval_slots}
\end{figure}

\noindent \textbf{Balancing Prototype-Centroid Retrieval for Effective Query-Aware Reasoning.} 
Figure \ref{fig:retrieval_slots} examines how many prototypes and centroids should be retrieved (Eq. \ref{eq:topk}, \ref{eq:topm}) during query-aware long-term reasoning. Since retrieval proceeds in a coarse-to-fine manner, the choice of number of retrieval slots directly shapes both evidence quality and retrieval efficiency. Too few prototypes fail to cover the diverse high-level contexts present in long videos, reducing the chance that the correct event cluster is included in the search space explaining the weaker accuracy and ARS. Conversely, retrieving too many prototypes pulls in loosely related or noisy abstract clusters, weakening focus and increasing overhead. A similar trade-off appears for centroids: too few centroids  miss relevant fine-grained states needed for tasks like step recognition or causal trigger detection, while too many centroids introduce excessive detail, slowing evidence localization and hurting ARS. The heatmaps show that the best results occur at the balanced setting of $8$ prototypes and $24$ centroids, which provides adequate semantic coverage through prototypes while maintaining enough centroid diversity to capture subtle evidence shifts without increasing retrieval latency. This additionally indicates that although the long-term memory is diverse, only a selective subset is needed per query, highlighting the effectiveness of our query-aware retrieval strategy.

While K-means is an effective way of memory construction, latency is governed by number of retrieved tokens and the reasoning strategy decides whether those tokens yield meaningful performance. In Table \ref{tab:latency}, all 
settings retrieve, reason over K-means' based memory, but StreamReady’s hierarchical strategy preserves performance under small token budget and comparable latency.

\begin{table}[t!]
    \centering
    \caption{ \textbf{Token retrieval \& reasoning strategies under identical budget.}}
    \begin{tabular}{l|cccc}
    \hline
         \textbf{Method} & \textbf{Token} & \textbf{Acc.} &  \textbf{Lat.}\\
         \toprule
         Full memory & 108 & 56.4 & 8.7s  \\ 
         Single-stage  & 32 & 40.7 & {3.2s} \\
         \rowcolor{green!10}
         \textbf{Hierarchical}  & \textbf{32} & \textbf{56.6} & \textbf{3.1s} \\
         \bottomrule
    \end{tabular}
    \label{tab:latency}
\end{table}

\begin{table}[t!]
    \small
    \caption{\textbf{Effect of changing backbone of StreamReady.}}
    \resizebox{\linewidth}{!}{
    \begin{tabular}{l|cc|cc}
    \toprule
         \multirow{2}{*}{\textbf{Method}} & \multicolumn{2}{c|}{\textbf{ProReady-QA}} & {\textbf{StrmB}} & \textbf{Vid-MME}\\
         \cline{2-5}
         & Acc. & ARS & Acc. & Acc.\\
         \midrule
         Oryx-1.5-7B &53.9 {\footnotesize \textcolor{ForestGreen}{($\uparrow 14.4$)}} &0.64 {\footnotesize \textcolor{ForestGreen}{($\uparrow 0.38$)}}& 63.6 {\footnotesize \textcolor{ForestGreen}{($\uparrow 15.2$)}} & 64.3 {\footnotesize \textcolor{ForestGreen}{($\uparrow 3.8$)}}\\
         LLaVA-OV 7B &55.2 {\footnotesize \textcolor{ForestGreen}{($\uparrow 9.8$)}} &0.65 {\footnotesize \textcolor{ForestGreen}{($\uparrow 0.27$)}}& 60.3 {\footnotesize \textcolor{ForestGreen}{($\uparrow 12.9$)}}& 63.8 {\footnotesize \textcolor{ForestGreen}{($\uparrow 5.6$)}}\\
         \rowcolor{green!10}
         \textbf{Qwen-2-VL 7B} & \textbf{56.4} {\footnotesize \textcolor{ForestGreen}{($\uparrow 15$)}} & \textbf{0.69} {\footnotesize \textcolor{ForestGreen}{($\uparrow 0.35$)}} & \textbf{63.4} {\footnotesize \textcolor{ForestGreen}{($\uparrow 18.2$)}} & \textbf{65.8} {\footnotesize \textcolor{ForestGreen}{($\uparrow 2.5$)}}\\
         \bottomrule
    \end{tabular}
    }
    \label{tab:backbone_ablation}
\end{table}
\noindent \textbf{Importance of Relevance-Gated Contextual Reasoning.} While we perform contextual reasoning through a similarity-based gating and relevance-filtering mechanism over the contextual memory bank, we also evaluate a variant that attends to all prior QA pairs without any filtering.
This naive design reduced accuracy by roughly $4-5$\% on StreamingBench, showing that blindly incorporating the entire QA history introduces noise from unrelated turns.  
In long streaming videos, such unfiltered cross-attention forces the model to process irrelevant reasoning traces, weakening answer correctness and disrupting temporal alignment by pulling retrieval toward outdated or mismatched evidence. In contrast, our similarity-guided selection ensures that only semantically aligned prior QA pairs influence the current reasoning, allowing contextual memory to serve as a focused and beneficial signal rather than a source of distraction.

\noindent \textbf{Robustness to Different Backbones.} Table \ref{tab:backbone_ablation} shows that StreamReady remains highly robust to the choice of 7B-scale MLLM backbone, consistently achieving large improvements on streaming benchmarks such as ProReady-QA and StreamingBench. This consistency indicates that the performance gains stem from our framework’s design rather than from any specific backbone architecture. By contrast, the improvements on VideoMME are noticeably smaller across all backbones. This difference highlights a key distinction: existing large MLLMs are already competitive on offline, single-query video reasoning, but they lack the temporal grounding, evidence tracking, and real-time retrieval capabilities required for true streaming understanding. They also lack mechanisms for answer timing based on evidence, which our framework introduces, reflected in the substantial ARS gains seen on ProReady-QA.

\begin{table}[t!]
    \centering
    \caption{\textbf{Effect of Readiness Supervision} using ProReady-QA. \ddag\ denotes original setup}
    \small
    \begin{tabular}{l|cc}
    \toprule
         \textbf{Methods and Timing Supervision} & \textbf{Acc.} & \textbf{ARS} \\
         \midrule
          StreamBridge \cite{wang2025streambridge} w/ BCE \ddag & 53.1 & 0.60 \\
          StreamBridge \cite{wang2025streambridge} w/ Contrastive & 53.2 & 0.63 \\
         \midrule
         StreamReady w/o timing sup. & 56.3 & 0.58\\
         StreamReady w/ BCE & 56.4 & 0.65\\
         \textbf{StreamReady} w/ \cellcolor{green!10}\textbf{Contrastive \ddag} & \cellcolor{green!10}\textbf{56.4} & \cellcolor{green!10}\textbf{0.69} \\
         \bottomrule         
    \end{tabular}
    \label{tab:timingloss}
\end{table}

\begin{figure}
    \centering
    \includegraphics[width=\linewidth]{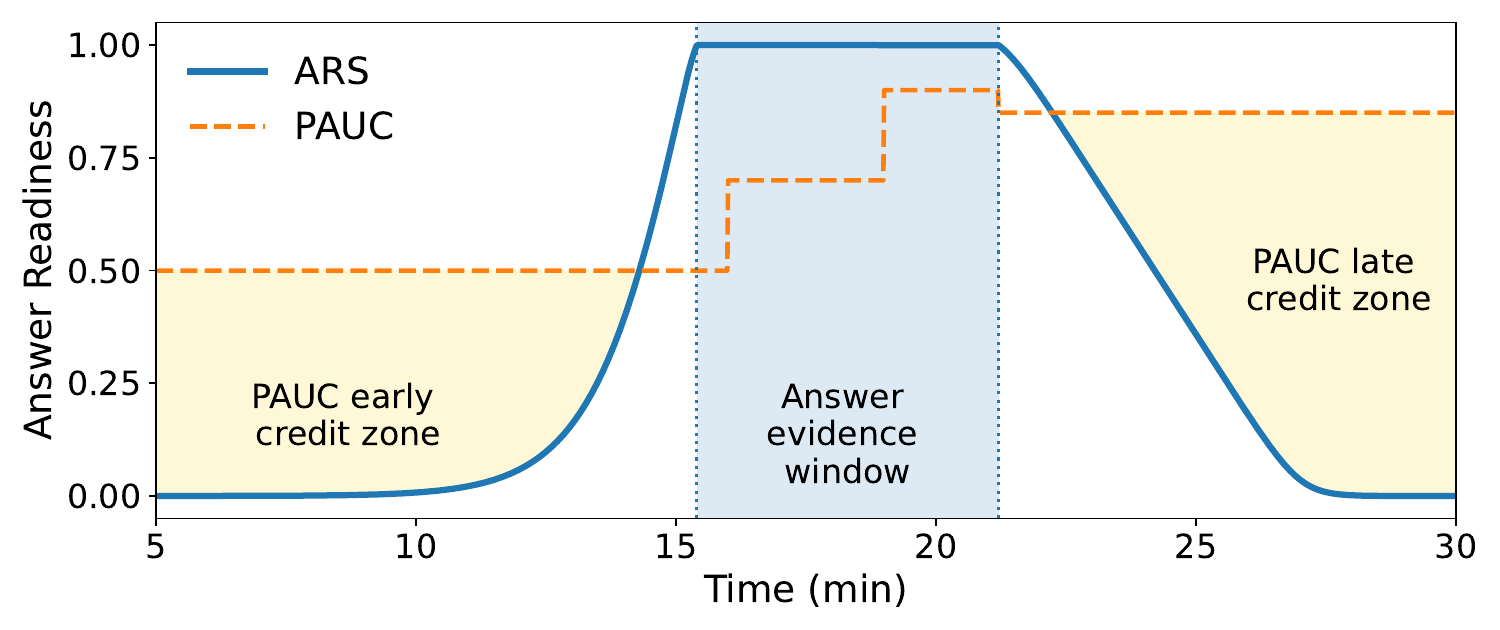}
    \caption{\textbf{Temporal characteristics of ARS} in relation to PAUC \cite{wang2025proactivevideoqacomprehensivebenchmarkevaluating}.}
    \label{fig:prs_pauc}
\end{figure}

\begin{figure*}
    \centering
    \includegraphics[width=\linewidth]{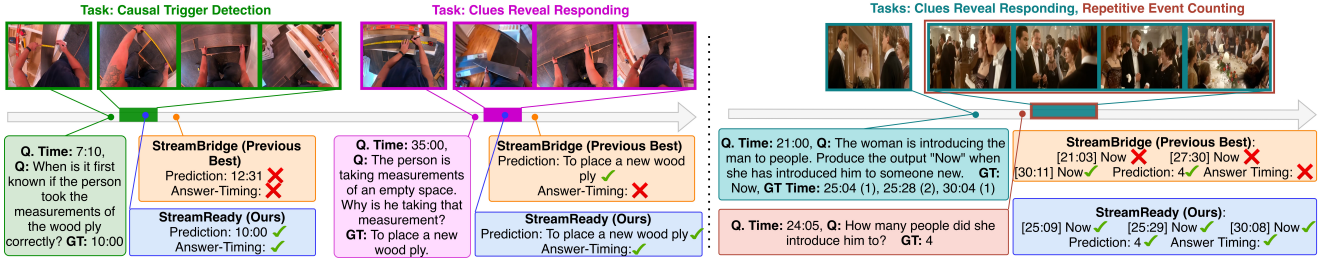}
    \caption{\textbf{Qualitative analysis} of readiness-aware streaming understanding on ProReady-QA. StreamReady shows superior performance by consistently giving accurate and on-time answers.}
    \label{fig:qual_sup}
\end{figure*}

\noindent \textbf{Impact of Timing Supervision and Ensuring Fair Comparison.} Table \ref{tab:timingloss} shows how different timing signal supervision strategies affect readiness modeling. For StreamReady, moving from no timing signal to BCE-based supervision yields only a modest improvement in ARS, since BCE provides a coarse binary view of readiness and does not meaningfully shape temporal behavior. Our contrastive supervision produces the largest ARS gain by learning an evidence driven readiness window, and trains the \texttt{<RDY>} pathway to recognize when evidence becomes sufficient. Importantly, accuracy remains effectively unchanged across the three variants, indicating that timing supervision influences readiness rather than correctness.

To ensure fairness when comparing with non–timing-aware methods, all offline baselines are triggered using the same confidence threshold as our readiness mechanism. Among existing online approaches, only StreamBridge \cite{wang2025streambridge} includes an explicit activation module; however, its design decouples readiness from reasoning by using a separate auxiliary network that does not access the same evidence or memory as the reasoning module. As a result, its readiness predictions are not grounded in the retrieved visual context. Even when trained with our contrastive timing objective (Eq. \ref{eq:totloss}), StreamBridge shows limited ARS improvement and still underperforms our method while incurring higher compute and latency overhead. This shows that StreamReady’s gains stem primarily from its unified evidence-aware reasoning and readiness mechanism, rather than its the timing supervision.

Importantly, the pseudo-labels used for training the readiness signal to provide timing supervision are derived from similarity between the \emph{reasoning module’s answer representation and visual memory}, rather than direct query–memory matching, thereby reflecting the outcome of reasoning. As a result, supervision reflects the outcome of reasoning rather than low-level visual cues. 
These labels
supervise the readiness pathway to only determine \emph{when} sufficient evidence has accumulated,
while the reasoning pathway determines \emph{what} constitutes valid evidence.
Since training datasets lack ground-truth evidence windows, 
we validate pseudo-labels post hoc on ProReady-QA, observing strong alignment with annotated evidence (mean temporal IoU $= 0.87$ over 100 samples).

\noindent \textbf{Robustness to Dense Tasks.} Although evaluated in QA settings, the readiness mechanism is not tied to VQA outputs or task-specific supervision. It is trained with task-agnostic pseudo-labels of evidence sufficiency (\S \ref{subsec:ans_ready}), and remains effective even without it (Table \ref{tab:timingloss}). Results on ET-Bench (dense video captioning, step localization and captioning, temporal video grounding), demonstrate consistent gains on open-ended generation/dense tasks (Table \ref{tab:etbench}), indicating the readiness mechanism's task-agnostic generalization.

\begin{table}[t!]
    \centering
    \caption{\textbf{Dense task performance (F1-score) comparison on ET-Bench.}}
    \begin{tabular}{l|ccc}
    \hline
         \textbf{Method} & \textbf{DVC} & \textbf{SLC} & \textbf{TVG}\\
         \toprule
         Qwen2-VL & 22.6 & 13.2 & 25.3\\
         Dispider & 33.8 & 18.8 & 36.1\\
         StreamBridge & 38.3 & 22.6 & 34.3\\
         \rowcolor{green!10}
         \textbf{StreamReady} & \textbf{43.4} & \textbf{24.1} & \textbf{36.8}\\
         \bottomrule
    \end{tabular}
    \label{tab:etbench}
\end{table}

\begin{figure}[t!]
    \centering
    \includegraphics[width=.7\linewidth]{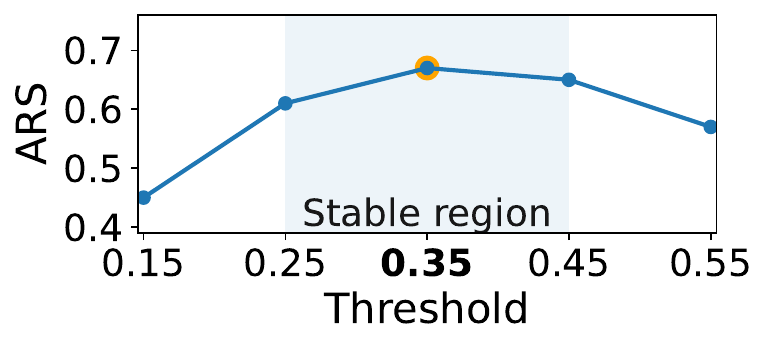}
    \caption{\textbf{Effect of readiness threshold (ProReady-QA).}}
    \label{fig:thr_sweep}
\end{figure}

\noindent \textbf{Effect of Readiness Threshold.}
Figure \ref{fig:thr_sweep} shows ARS stability over a broad range 
of readiness threshold 
around the default value ($0.35$), degrading only at extremes due to premature or delayed triggering, indicating model robustness to threshold choice.  

\noindent \textbf{Probing Unanswerability.} While a fully general readiness model should also identify when a question is unanswerable, the primary focus of this work is to study \emph{when} a model should answer, given that an answer exists, which is consistent with existing streaming benchmarks and allows us to analyze response timing. That said, StreamReady couples readiness with reasoning over accumulated evidence: if a question is never answerable, the memory never contains sufficient query-aware evidence, and the model’s readiness signal remains low throughout the stream. To validate this, we probe unanswerability with 30 counterfactual questions across random ProReady-QA videos, and find that StreamReady’s readiness score remains low throughout (avg. $0.21$), never triggering a response. This reflects its evidence-coupled readiness design that suppresses readiness in the absence of sufficient evidence.
ARS also naturally handles such cases ($\tau, t_s, t_e = 0$ Eq. \ref{eq:ep}, \ref{eq:lp}), yielding near-zero effective accuracy, even if an answer were produced.
Explicit modeling of unanswerability remains a promising future direction.

\subsection{ARS Metric Behavior Analysis}

\noindent \textbf{Metric Rationale.}
ARS is designed to reflect how well a model’s answers align with the evolving availability of visual evidence, so each component of the formulation is chosen to make timing behavior comparable, stable, and interpretable across diverse videos. The median evidence duration ($\tau$) provides a consistent temporal scale that prevents overly harsh penalties in videos containing short evidence windows and overly lenient ones in long windows.
The factor of $2$ in EP (Eq. \ref{eq:ep}) ensures that answers given exactly at the evidence onset ($t_a=t_s$) receive full credit, avoiding the midpoint behavior of the sigmoid function that would otherwise mark an on-time answer as partially early with $0.5$ penalty. 
Softmin and softmax operators enforce smooth transitions at window boundaries, preventing abrupt scoring jumps that could unfairly reward or penalize models near start or end of window. 

ARS also resolves several practical edge cases. When an answer is given far before the evidence window, EP saturates naturally toward zero, indicating that no part of the response is supported by the visual stream. When $t_s < t_a < t_e$, both EP and LP remain at 1, reflecting that the answer is fully grounded. After delays beyond $t_e$, LP decays gradually rather than collapsing, which avoids brittle behavior in tasks where evidence dissolves slowly.
The formulation also behaves sensibly under atypical annotation cases: if noisy annotations ever produce $t_s > t_e$, the monotonicity of the penalties ensures that ARS does not produce contradictory or negative values. Altogether, the metric components work jointly to provide a consistent and interpretable estimate of answer readiness that aligns with how visual evidence unfolds in streaming video scenarios.

\noindent \textbf{Temporal Characteristics of ARS.}
Figure \ref{fig:prs_pauc} contrasts ARS with the timing formulation used in PAUC \cite{wang2025proactivevideoqacomprehensivebenchmarkevaluating} to highlight how different scoring schemes treat answer timing in streaming settings. PAUC assigns temporal credit using a fixed baseline before the evidence window and then propagates the model’s last correctness score after the window ends. This makes it less responsive to early or late answers and less reflective of gradual changes in when evidence becomes supportive. ARS instead models readiness as a continuous, asymmetric curve that rises only when visual evidence becomes valid and declines as that evidence fades. This produces a smoother alignment between the model’s response timing and the underlying visual support. The figure highlights these distinctions in temporal behavior and helps clarify why ARS provides a natural fit for evaluating evidence-based readiness in long streaming videos. 

\noindent \textbf{Extending to Datasets without Annotated Evidence Window.} Although ARS is defined using annotated evidence windows, its underlying requirement is evidence sufficiency, rather than precise temporal boundaries. While explicit windows offer the most reliable signal, sufficiency can also be approximated using native temporal annotations (e.g. action/scene boundary, timestamp) or model-driven cues like answer stabilization or cross-model agreement, enabling readiness-aware evaluation even in benchmarks without annotated evidence windows. 

\subsection{Qualitative Analysis}
In Figure \ref{fig:qual_sup}, we present a qualitative comparison of StreamReady and StreamBridge on the three tasks (CTD, CRR, REC) of the ProReady-QA benchmark. Across all cases, StreamReady delivers accurate and timely answers because its readiness signal is tied directly to the same evidence used for reasoning through the hierarchical visual memory. In contrast, StreamBridge often misfires because its activation module is decoupled from reasoning, which leads to mistimed outputs, ARS penalties, and answers that drift from the actual visual evidence.

In the CTD example on the \textit{left}, StreamBridge activates only after the causal cue has passed and therefore reasons over the wrong visual moment, producing an incorrect answer and a late penalty. StreamReady detects the cue when it first appears and responds correctly within the evidence window.
In the CRR example at the \textit{center}, StreamBridge performs accurate prediction but triggers after the evidence window ends, harming its ARS and effective accuracy. StreamReady tracks the gradual buildup of clues and answers as soon as the evidence becomes sufficient.

The combined REC+CRR example on the \textit{right} shows the most significant failure for StreamBridge. It answers the CRR question prematurely at question time, hallucinating a trigger that does not yet exist and incorrectly counting two individuals during that time. Later, it misses a real event within the evidence window, but still outputs the correct count, producing an answer that is not grounded in the visual sequence; thus incurring ARS penalty and reduced effective accuracy. StreamReady avoids this issue by recognizing that the CRR question is not answerable at question time, waiting for the true clue, and counting only events supported by evidence.

\begin{figure}[t!]
    \centering
    \includegraphics[width=.96\linewidth]{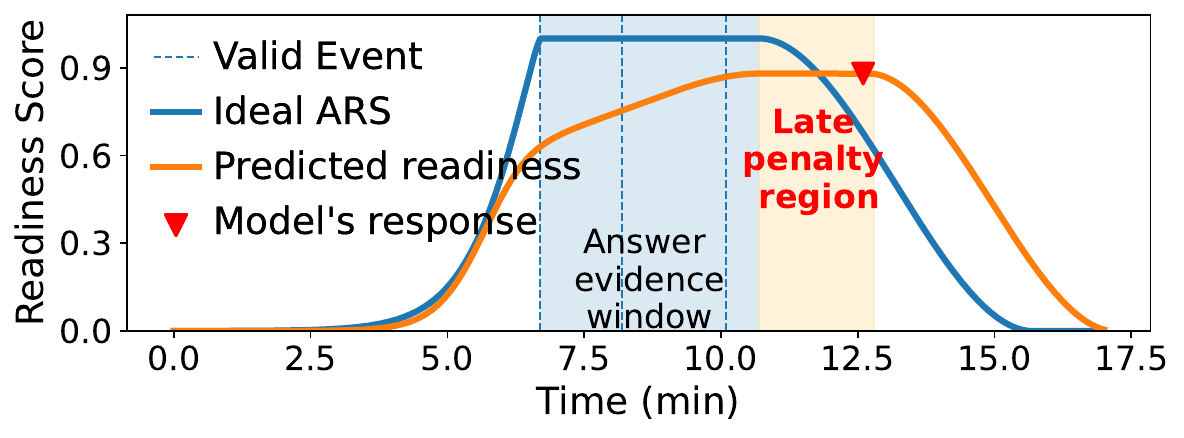}
    \caption{\textbf{Readiness failure in counting task.}}
    \label{fig:readiness}
\end{figure} 

On the other hand, Figure \ref{fig:readiness} shows a failure case in counting, where the readiness signal remains high beyond evidence window, as the model waits to confirm that no additional relevant event occurs. This reflects the inherent ambiguity of such tasks, leading to delayed responses and a late penalty under ARS. 

Overall, these examples illustrate that StreamReady’s combined retrieval, reasoning, and readiness design enables precise evidence tracking and on-time answering in timing-sensitive streaming tasks.

\section{ProReady-QA Generation Pipeline.}
\label{sec:generation_supp}
ProReady-QA is constructed from long-form Ego4D and MovieNet videos using a semi-automatic pipeline following \cite{zhang2024flashvstream, li2025ovobench} to generate proactive QA pairs and answer evidence window annotations. 
Although the dataset contains fewer source videos than some earlier benchmarks, each video is significantly longer and paired with richer temporal annotations, enabling readiness-aware evaluation under ARS.

\noindent \textbf{Dense Captioning.} Each video is divided into 30-second segments, from which 8 uniformly sampled frames are processed by Qwen-2-VL to produce dense captions describing actions, objects, interactions, and spatial context, along with segment timestamps. 

\noindent \textbf{Summarization.} We then aggregate dense captions over several-minute intervals and summarize them using an Qwen-2, preserving key entities, actions, causal structure, and temporal ordering while avoiding redundancy.

\noindent \textbf{Multi-Turn Proactive QA Generation.} To construct future-dependent QA, we take consecutive scene summaries and prompt the VLM to generate a question that a viewer might ask at the end of Scene~A and whose answer only becomes knowable in Scene~B. This enforces strict temporal ordering and ensures that every question is fundamentally future-dependent. From this pool of valid proactive QA, we then construct multi-turn dialogues for a subset of segments. The VLM is prompted to propose follow-up questions that explicitly reference earlier turns; either the immediately preceding question or one from farther back, while still requiring new future evidence to answer. These multi-turn chains may span different task types. 

\noindent \textbf{Answer Evidence Timestamp Annotation.} 
We extract coarse candidate evidence spans using multimodal cues from native annotations (actions, interactions, scene changes, or subtitle segments) to anchor where the answer is likely revealed. Within this restricted window, a frame-wise VLM check identifies the earliest frame where the answer becomes inferable and the latest frame where the supporting cue remains valid. We refine these boundaries using dwell-based smoothing and task-specific rules (e.g., step transitions for SSR, clue sufficiency for CRR, count completion for REC, goal realization for GSD, causal effect visibility for CTD). This process yields evidence windows appropriate for timing-aware evaluation through ARS.

\noindent \textbf{Human Refinement.} After the automatic pipeline, annotators review each QA pair and its proposed evidence window. They verify that the answer is not inferable before $t_s$, that the supporting cue truly disappears at $t_e$, and that the QA instance is strictly future-dependent. Ambiguous or trivial cases, hallucinated cues, or QA pairs whose evidence does not align with the timestamps are corrected or removed, with replacement added when needed; especially for complex tasks such as CTD and CRR.

\noindent \textbf{Quality Control.} We assess annotation reliability on 100 ProReady-QA samples, where two annotators achieve high agreement on evidence windows (mean temporal IoU = $0.85$), with model-based answer verifiability rising sharply within these windows ($0.35$ to $0.82$), confirming annotation reliability.

\section{Implementation Details}
\label{sec:imp_sup}
Table \ref{tab:train_data} shows the training data used for fine-tuning the reasoning module (\S \ref{subsec:qf}) and the readiness mechanism (\S \ref{subsec:ans_ready}), while the visual encoder and language decoder remain frozen.  The trainable parameters are trained for $5$ epochs with a learning rate of $2e-5$ with a cosine annealing scheduler
and AdamW optimizer ($[0.9, 0.999]$). We use $\alpha = 0.985$ for EMA decay factor in Eq. \ref{eqn:mv2}, \ref{eqn:mv3}. $\lambda_{reg} = 0.1$ in Eq. \ref{eq:totloss}. All videos are sampled at 1 FPS following standard protocol of streaming video understanding and input video frames are resized to $224\times224$.

\begin{table}[t!]
    \centering
    \caption{\textbf{Dataset Statistics} for training the reasoning module and readiness mechanism}
    \begin{tabular}{l|lc}
        \toprule
         Task &  Dataset & \#QA Pairs \\
         \midrule
         \multirow{4}{*}{VQA} & MSRVTT-QA \cite{xu2017video} & 10k \\
         & MSVD-QA \cite{xu2017video}& 10k\\
         & ActivityNet-QA \cite{yu2019activitynet}  & 32k\\
         & MovieChat-1k \cite{song2024moviechat} & 10k\\

         \bottomrule
    \end{tabular}
    \label{tab:train_data}
\end{table}

\section{Future Work}
\label{sec:future}
While StreamReady and ProReady-QA advance readiness-aware streaming video understanding, several promising directions remain open for future exploration. Extending the framework beyond a single-view setting to multi-camera or multi-agent environments would allow reasoning over parallel visual threads, where evidence may appear asynchronously across different viewpoints. Enabling multimodal streaming input (e.g. audio) could further enhance readiness estimation in scenes where critical cues are partially non-visual. Another natural extension is to explore readiness-aware behavior in interactive or embodied agent settings, where a model can decide not only {when} to answer but also {when} to request more information, ask clarifying questions, or seek an alternative viewpoint. Such capabilities are increasingly important for agentic frameworks, and incorporating readiness into these systems may help them maintain evidence-grounded decision-making in long streaming contexts.

\end{document}